\documentclass[10pt,twocolumn,letterpaper]{article}

\usepackage[pagenumbers]{cvpr} 

\usepackage{multirow}
\usepackage{stfloats}
\usepackage{wrapfig}
\usepackage[dvipsnames]{xcolor}

\definecolor{cvprblue}{rgb}{0.21,0.49,0.74}
\usepackage[pagebackref,breaklinks,colorlinks,citecolor=cvprblue]{hyperref}

\title{Control Color: Multimodal Diffusion-based Interactive Image Colorization}

\author{Zhexin Liang \quad Zhaochen Li \quad Shangchen Zhou \quad Chongyi Li \quad Chen Change Loy\\
S-Lab, Nanyang Technological University\\
{\tt\small \{zliang008, c170086, s200094, ccloy\}@ntu.edu.sg, lichongyi25@gmail.com}\\
{\tt\small \url{https://zhexinliang.github.io/Control_Color}}
}

\begin{document}

\twocolumn[{%
   \renewcommand\twocolumn[1][]{#1}%
   \maketitle
   \begin{center}
    \centering
    \vspace{-4mm}
    \includegraphics[width=\linewidth]{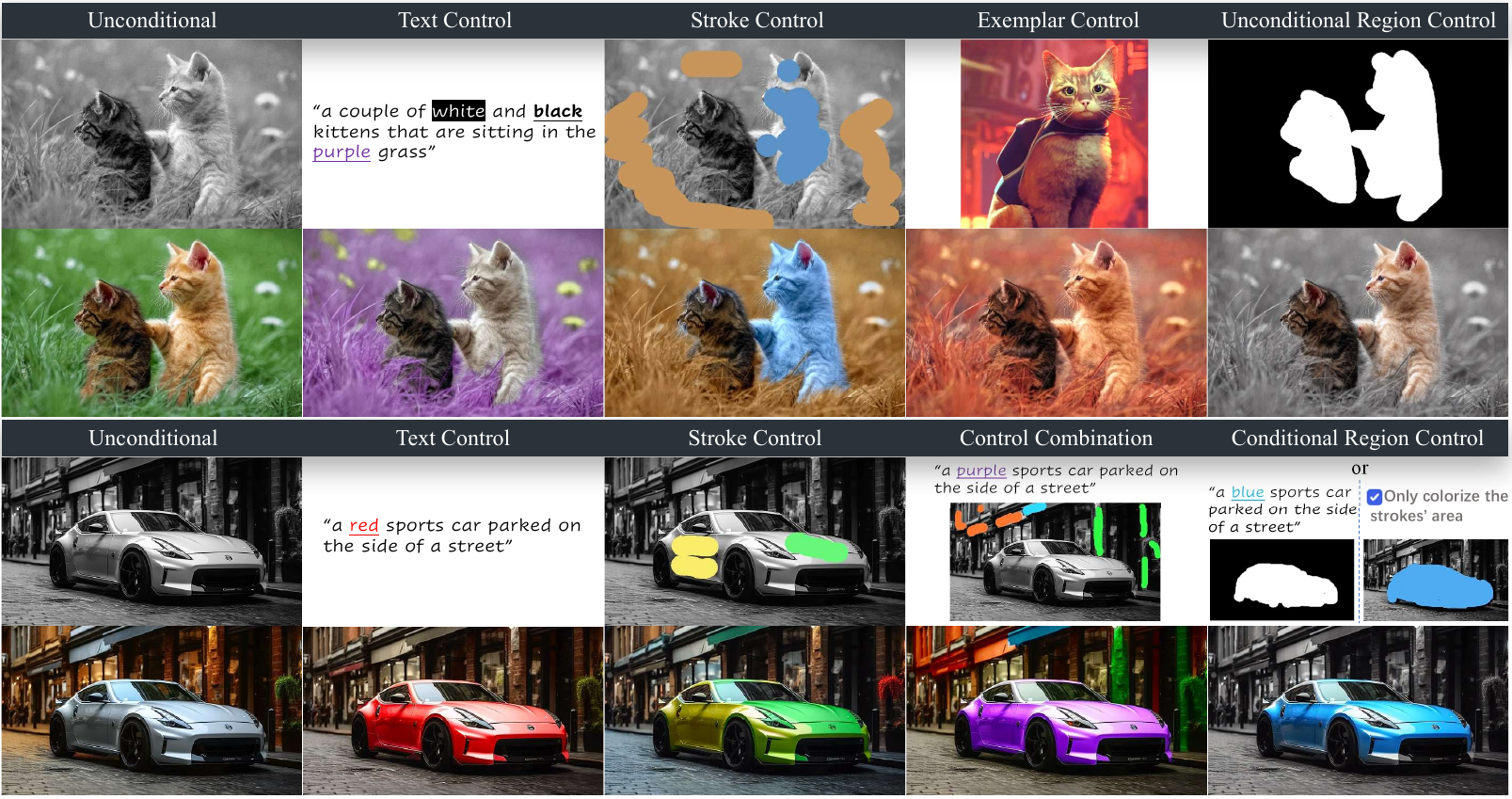}%
    \vspace{-2mm}
    \captionof{figure}{The proposed CtrlColor achieves highly controllable image colorization, offering users a simple and intuitive means to colorize images according to their specific preferences. Our method supports both unconditional and conditional colorization, including options such as text, stroke, and exemplar image, as well as allows for any combination of them. Utilizing strokes as masks, our method facilitates selective image editing effortlessly. Our method also supports highly flexible iterative image editing, empowering users to finely tune specific details during the colorization process.
 }
 \vspace{1mm}
    \label{fig:teaser}
   \end{center}%
}]
\maketitle
\begin{abstract}
\vspace{-1mm}
Despite the existence of numerous colorization methods, several limitations still exist, such as lack of user interaction, inflexibility in local colorization, unnatural color rendering, insufficient color variation, and color overflow. 
To solve these issues, we introduce Control Color (CtrlColor), a multi-modal colorization method that leverages the pre-trained Stable Diffusion (SD) model, offering promising capabilities in highly controllable interactive image colorization.
While several diffusion-based methods have been proposed, supporting colorization in multiple modalities remains non-trivial.
In this study, we aim to tackle both unconditional and conditional image colorization (text prompts, strokes, exemplars) and address color overflow and incorrect color within a unified framework.
Specifically, we present an effective way to encode user strokes to enable precise local color manipulation and employ a practical way to constrain the color distribution similar to exemplars. Apart from accepting text prompts as conditions, these designs add versatility to our approach.
We also introduce a novel module based on self-attention and a content-guided deformable autoencoder to address the long-standing issues of color overflow and inaccurate coloring.
Extensive comparisons show that our model outperforms state-of-the-art image colorization methods both qualitatively and quantitatively. 
\end{abstract}
    
\vspace{-4mm}
\section{Introduction}
\label{sec:intro}

\begin{figure}[!t]
    \begin{center}
    \vspace{-2mm}
    \includegraphics[width=\columnwidth]
    {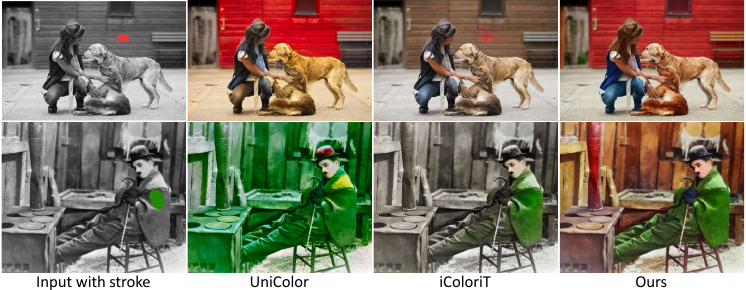}
    \end{center}
    \vspace{-6mm}
    \caption{Visual comparisons on stroke-based colorization.}
    \label{fig:stroke0}
    \vspace{-6mm}
\end{figure}

Image colorization aims at colorizing grayscale images, which enhances the visual appeal in various domains such as historical footage.
Manual colorization is a labor-intensive process that heavily relies on the preferences, experience, imagination, and laborious effort of colorists.
Although many automatic colorization methods \cite{cheng2015deep,zhang2016colorful,zhao2020pixelated,su2020instance,vitoria2020chromagan,wu2021towards,kumar2021colorization,kim2022bigcolor,huang2022unicolor,xia2022disentangled,saharia2022palette,yun2023icolorit, xiao2019interactive} have been developed,  they still exhibit certain drawbacks. These include limited color richness, color overflow, color distortions, and incomplete coloring in certain areas.
Furthermore,  methods that rely on user-provided conditions lack flexibility and precision.
For example, these kinds of methods do not allow for selective colorization of specific regions or the use of strokes to apply color to particular objects. 
Recent image colorization techniques, such as those in UniColor~\cite{huang2022unicolor} and iColoriT~\cite{yun2023icolorit}, offer stroke-based color control but are constrained to minor square hint points within object boundaries. This necessitates more exact hint placement, and the methods may struggle with significant color overflow when colorizing small areas or cannot correctly colorize images based on stroke's color, as depicted in Fig.~\ref{fig:stroke0}. In addition, these approaches do not facilitate region colorization due to their simplistic designs, which merely extend the hint point colors to adjacent regions.

Stable Diffusion (SD)~\cite{rombach2022high} with ControlNet~\cite{zhang2023adding}, which was originally proposed for conditional image generation, appears to be a viable solution for image colorization.
Although it is capable of generating diverse and high-quality images with flexible conditions, it does not explore how to integrate multiple conditions within a unified framework, relying instead on a single control such as Canny edges or segmentation maps. Furthermore, due to the high sparsity of conditions and the inherent randomness in the diffusion process, it tends to generate results with low fidelity, making it unsuitable for our multimodal image colorization task.

To address the challenges mentioned above, we propose a novel multi-modal diffusion-based colorization framework, called \text{CtrlColor}. This framework aims to unify various colorization tasks including unconditional, prompt-, stroke-, and exemplar-based image colorization within a single framework. 
CtrlColor leverages rich priors encapsulated in the latent diffusion model (\ie, SD~\cite{rombach2022high}) that is trained on a large-scale image dataset.
This enables our method to obtain significantly superior results compared to previous approaches in terms of color richness and diversity. 
To address the color overflow and incorrect color issues, we introduce self-attention guidance and content-guided deformable autoencoder in our framework. Without any training, self-attention guidance is added in the inference process to address the small color overflow problem by blurring the out-of-distribution attention area and regenerating the blurred area with color distributions that are more similar and harmonious to their surroundings. To handle more complex and heavy color artifacts caused by the low fidelity characteristics of the diffusion model, we introduce deformable convolution layers~\cite{dai2017deformable} into the decoder of SD autoencoder. Guided by the content of the input images, these layers constrain regions of the deformation to align the generated colors to the input textures, consequently reducing issues of low color fidelity,~\ie, color overflow and incorrect color.

Moreover, we present a new method to enable stroke-based colorization. In particular, we incorporate combinations of a hint point map, a binary mask, and an input grayscale image into the SD model. The hint points are located using strokes. To indicate the positions of the hint points explicitly, we introduce a binary mask derived from the hint point map.
By encoding the hint points into the latent space during the denoising process, we gain control over the denoised color distribution. This allows users to flexibly modify the color of an image using arbitrary strokes, enabling modifications on specific local regions. 
We leave the details of prompt-based and exemplar-based colorization to the methodology section.

In this study, we present the diverse applications of our method through a preview of our results, as shown in Fig.~\ref{fig:teaser}. 
Our approach offers a versatile and effective solution for implementing highly controllable image colorization, achieving state-of-the-art performance in terms of color richness, stability, and visual quality for image colorization.
We also provide a video demo to showcase an interactive interface that demonstrates the multi-modal colorization and local controllability of our method.

To summarize, our main contributions are as follows: 1) We propose a novel diffusion model-based framework, CtrlColor, which enables highly controllable multi-modal colorization. Our framework supports prompt-based, stroke-based, exemplar-based, and a combination of these conditions for local and global colorization.  
2) Our approach addresses the color overflow problem by employing training-free self-attention guidance and the learned content-guided deformable autoencoder. 
3) We introduce a new approach to add user strokes to control the color locally and precisely by encoding the strokes' position and color into the diffusion process. 
4) Besides overcoming the low fidelity of the pre-trained SD model, our method leverages its advantages, producing colored images that exhibit a richer variation in color compared to previous methods.

\section{Related Work}
\label{sec:related_work}

\noindent
\textbf{Unconditional Colorization.} Unconditional colorization aims to colorize grayscale images automatically. 
Early attempts formulate colorization as either a regression~\cite{cheng2015deep} or a classification task~\cite{zhang2016colorful}.
To further guide colorization with semantic information, class labels~\cite{kim2022bigcolor}, semantic segmentation maps~\cite{zhao2020pixelated}, or/and instance bounding box~\cite{su2020instance} are incorporated into colorization networks. 
The recent Generative Adversarial Network (GAN)~\cite{goodfellow2020generative} and Transformer~\cite{vaswani2017attention} have also shown promising results in colorization. 
DeOldify~\cite{antic2019deoldify} and ChromaGan~\cite{vitoria2020chromagan} directly optimize the GAN-based networks, whereas ~\citet{wu2021towards} and ~\citet{kim2022bigcolor} exploit the generative prior of the pre-trained GANs. 
Benefiting from its long-range receptive field, Transformer architecture is used to auto-regressively predict color tokens either in pixel space~\cite{kumar2021colorization} or in the latent space~\cite{huang2022unicolor,xia2022disentangled, kang2023ddcolor}. 
More recently, Palette~\cite{saharia2022palette} trains a diffusion model from scratch conditioned on grayscale images to generate diverse colors. However, the knowledge of color hint is still restricted to the limited colorization training data, hindering its ability to generate more vivid colors.
\begin{figure*}[!t]
\vspace{-6mm}
    \centering
    \includegraphics[width=1.0\linewidth]{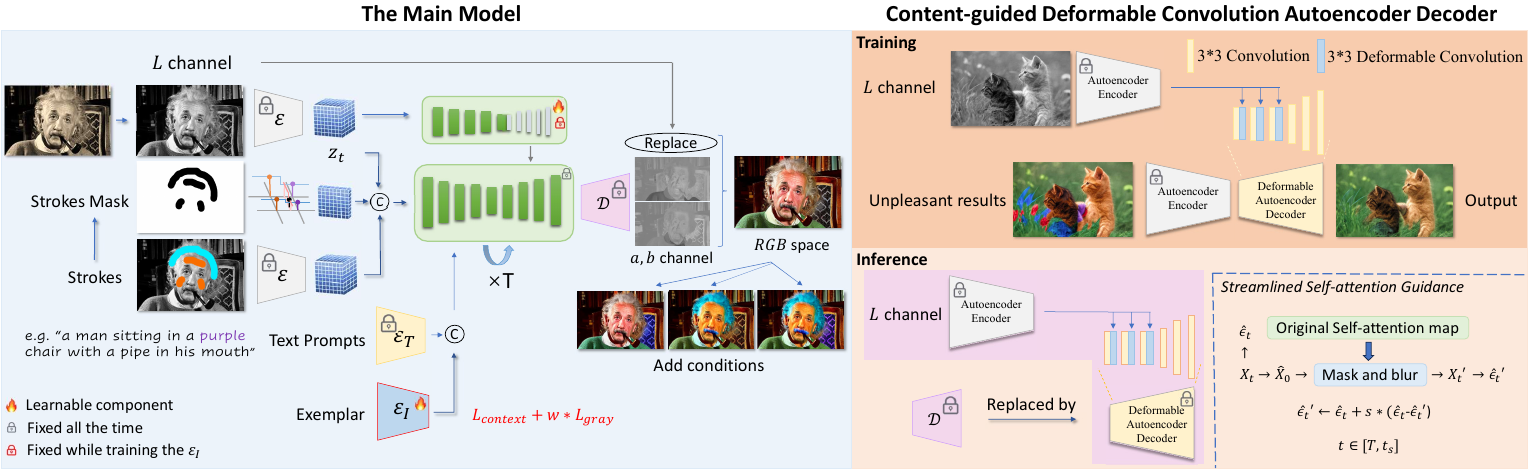}%
    \vspace{-3mm}
    \caption{\textbf{Left}: The main structure of our CtrlColor Model achieves multi-modal controllable colorization by blending controls from diverse components. \textbf{Right}: To manage large color overflow and inaccurate color regions, we integrate content-guided deformable convolution layers into the autoencoder's decoder. These layers restrict deformed color regions to align with nearby colors. Additionally, refined self-attention guidance is employed during inference to blur small overflow areas by referencing the surrounding color distribution. This process aims to smooth the color distribution, effectively addressing the issue of small color overflow.}
    \label{fig:img_pipeline}
    \vspace{-4mm}
\end{figure*}

\noindent
\textbf{Conditional Colorization.} Conditional colorization adds user controls to the colorization process. Based on input modality, it can be categorized into three types: stroke-based colorization, exemplar-based colorization, and prompt-based colorization.
Prior stroke-based colorization works rely on similarity metrics~\cite{levin2004colorization, luan2007natural} such as spatial offset or learned through neural networks~\cite{endo2016deepprop} to propagate local color hints to the whole image. 
Recent  methods use end-to-end propagation frameworks~\cite{huang2022unicolor, yun2023icolorit, xiao2019interactive}. 
Exemplar-based colorization works mainly employ a pre-trained network for feature matching between the input grayscale image and the exemplar image~\cite{he2018deep, zhang2019deep}. 
\citet{huang2022unicolor} further regulates the matching matrix by converting only cells with high matching confidence to hint points.
Prompt-based colorization takes text descriptions as guidance to assist colorization. One direction revolves around fusing text and image features. Feature-wise affine transformation~\cite{Manjunatha:Iyyer-ColorLanguage} and recurrent attentive model~\cite{chen2018language} have been proposed. 
Some other studies focus on decoupling color and object feature space to avoid color-object mismatch and color-object coupling~\cite{weng2022code}.
More recent works~\cite{chang2023cad, zabari2023diffusing} use diffusion prior to achieve prompt control, but the output size of L-CAD~\cite{chang2023cad} is limited and the results of Diffusing Colors~\cite{zabari2023diffusing} suffer from severe color overflow.
UniColor \cite{huang2022unicolor}, leveraging Transformer~\cite{vaswani2017attention} and VQGAN~\cite{esser2021taming}, is a seminal work unifying all three types of conditions by representing them as hint points across fixed-sized cell regions. However, one cell region consists of many pixels, possibly with diverse colors. Simply representing them with one color may lead to undesired results like color bleeding. 
In contrast, our approach manipulates different conditions in latent space via various encoding approaches and adopts a pixel-level binary mask for strokes, achieving highly precise and flexible control.

\noindent
\textbf{Diffusion Models for Image Generation.}
The diffusion probabilistic model has shown successful results in image generation~\cite{dhariwal2021diffusion, ho2020denoising, song2020denoising}, which is extended to latent space~\cite{rombach2022high} and text conditioned generation~\cite{saharia2022photorealistic, nichol2021glide}.
GLIDE~\cite{nichol2021glide} and Imagen~\cite{saharia2022photorealistic} encode text inputs into latent vectors using pre-trained vision or/and language models like CLIP~\cite{radford2021learning} and T5~\cite{raffel2020exploring}. 
To further enhance control over diffusion models, several methods exploit input mask~\cite{avrahami2022blended, nichol2021glide}, encoded text features~\cite{hertz2022prompt, kawar2023imagic} and decompose image into basic control components~\cite{huang2023composer}.
Zhang et al. \cite{zhang2023adding} propose ControlNet to control diffusion model with task-specific conditions such as human key points and user scribbles by fine-tuning a ``trainable copy'' of any off-the-shelf diffusion model. However, training ControlNet na\"{i}vely with grayscale images as conditions generates images with noticeable differences compared to the input, particularly in textures and details, which is undesirable for colorization.

\section{Methodology}
\label{sec:method}

\subsection{Preliminary: LDM and ControlNet}
Diffusion Model consists of a forward and a backward process.
In the forward process, Gaussian noise is gradually added to the input image $\mathbf{x}_0$ in $T$ steps, producing a series of noisy samples $\mathbf{x}_1$, $\cdots$, $\mathbf{x}_T$. When $T$ is large enough, $\mathbf{x}_T$  approximates a Gaussian distribution.
In the backward process, a denoising neural network (typically in the form of U-Net) is used to recover $\mathbf{x}_0$ from $\mathbf{x}_T$, as shown in Eq.~\eqref{eq:x0}.
Specifically, in each time step $t$ during the diffusion reverse process, given the latent noised image code $X_t$, the diffusion model will output the predicted noise $\epsilon_t$, its variance $\sigma^2_t$, and the attention map $A_t$, and then we use the predicted noise to get $X_{t-1}$, \ie, the latent noised image code at step $t-1$. We use the $A_t$ to modify $\epsilon_t$ in our streamlined self-attention guidance.
\begin{equation}
    \hat{X_0}=\frac{X_t-\sqrt{1-\bar{\alpha}_t}\epsilon_t}{\sqrt{\bar{\alpha}_t}},
    \label{eq:x0}
\end{equation}
where $\bar{\alpha}_t=\prod^t_{i=1}(1-\sigma^2_t)$.

\noindent\textbf{Latent Diffusion Model (LDM).} LDM~\cite{rombach2022high} is proposed to run the diffusion process in latent space instead of pixel space, which saves computational costs. 
LDM introduces an autoencoder model for image reconstruction. The encoder $\mathcal{E}$ encodes the input image into a 2D latent space $\mathbf{z}$. The decoder $\mathcal{D}$ then reconstructs the image from $\mathbf{z}$. Thus, the overall LDM framework consists of a denoising network (U-Net) and an autoencoder. Our designs target these two components, as shown in Fig.~\ref{fig:img_pipeline}.
The overall loss can be expressed as:
\begin{equation}
    \mathcal{L}_{LDM} = \mathbb{E}_{
        \mathcal{E}(\mathbf{x}), 
        y, 
        \epsilon \sim \mathcal{N}(0, 1), 
        t
    } 
    \Big[\|\epsilon - \epsilon_\theta(z_t, t)\|^2_2 \Big],
\end{equation}
where $t$ denotes the diffusion step, while $\epsilon_\theta$ is the denoising network. 

\noindent\textbf{ControlNet.} ControlNet~\cite{zhang2023adding}  uses ``trainable copy'' of LDM on task-specific datasets with an additional input condition $c_i$ apart from the text condition $y$, while the ``locked copy'' preserves the original weights of the pre-trained LDM. With these designs, ControlNet achieves more control over generated image content. 
The overall learning objective then becomes: 
\begin{equation}
    \mathcal{L} = \mathbb{E}_{
        \mathcal{E}(\mathbf{x}), 
        y, 
        c_i
        \epsilon \sim \mathcal{N}(0, 1), 
        t
    } 
    \Big[\|\epsilon - \epsilon_\theta(z_t, t, y, c_i)\|^2_2 \Big].
\end{equation}
By freezing the parameters of the pre-trained LDM to allow for efficient training, we successfully transfer the generative prior from LDM to image colorization tasks.

\subsection{Framework of Control Color}
As shown in Fig.~\ref{fig:img_pipeline}, our approach consists of two main components: 1) image colorization latent diffusion model to achieve multi-modal controls; 2) content-guided deformable autoencoder and streamlined self-attention guidance to handle color overflow and incorrect color issues. 
\vspace{-4mm}
\subsubsection{Unconditional Colorization}
\vspace{-1mm}
In Fig.~\ref{fig:img_pipeline} (left), we first convert the input image into $Lab$ space and get its $L$ channel. The $L$ channel is first encoded into the latent space using the autoencoder's encoder, then serves as an extra condition input to ControlNet. The input $RGB$ image is also encoded and used as input of the Stable Diffusion model. This operation allows the trained model to generate colorized images that closely resemble the structure of the input grayscale image.

Previous methods~\cite{zhang2016colorful, kang2023ddcolor} predict the $ab$ channels given the corresponding $L$. We employ a similar strategy, but only in the post-processing, to mitigate the small deformations introduced by the diffusion model while retaining the generated color characteristics. The output is reconstructed by our autoencoder's decoder and then converted from $RGB$ space to $Lab$ space. We replace the reconstructed $L$ channel with the original input $L$ channel and then convert the resulting $Lab$ image back to $RGB$  space to obtain the final output. This post-processing ensures the final output has the same content as the input. 

\begin{figure*}[!t]
    \centering
    \vspace{-8mm}
    \includegraphics[width=1.0\linewidth]
    {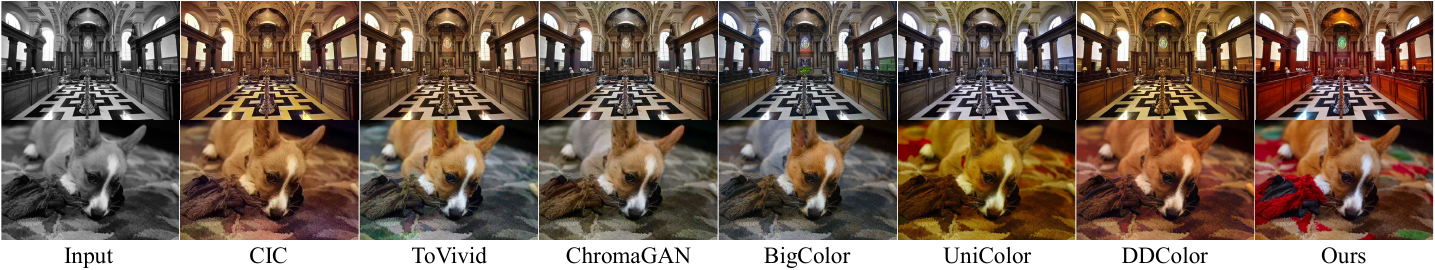}
    \vspace{-7mm}
    \caption{Qualitative comparison for unconditional image colorization. The first row of images is from the COCO-stuff dataset, and the second row comes from the ImageNet validation dataset. Our method generates more vivid and realistic colors with less color bleeding. (\textbf{Zoom-in for best view}) More comparisons are provided in the supplementary material.}
    \label{fig:uncond_compar}
    \vspace{-5mm}
\end{figure*}
\vspace{-4mm}
\subsubsection{Conditional Colorization}
\vspace{-1mm}
Based on the above unconditional colorization, we add multi-modal controls by the following designs. 

\noindent
\textbf{Prompt Control.}
Consistent with approaches adopted in prior research~\cite{zhang2023adding,rombach2022high}, our methodology involves initially encoding the text prompt using the CLIP~\cite{radford2021learning} text encoder. Subsequently, this encoded information is integrated into the intermediate layers of the U-Net architecture through a cross-attention layer, a process applied within both the ControlNet and Stable Diffusion model frameworks. Leveraging the cross-attention mechanism enables our model to effectively interpret and respond to text prompts, ensuring precise control over the generated content.

\noindent
\textbf{Stroke Control.}
The strokes are first directly overlaid on the $L$ channel images, denoted as hint images, so that we can get both the color and the position of the strokes. A binary mask is then derived from the hint image. The input image $I_i\in\mathbb{R}^{C\times H \times W}$ and the hint image $I_s\in\mathbb{R}^{C\times H \times W}$ are encoded into the latent space to get the input latent feature $z_i^{4\times\frac{H}{8}\times\frac{W}{8}}$ and the hint latent feature $z_s^{4\times\frac{H}{8}\times\frac{W}{8}}$, respectively. We apply nearest-neighbor downsampling to downsample the stroke mask to $z_m^{1\times\frac{H}{8}\times\frac{W}{8}}$. 
Then we concatenate them into a $\tilde{z}^{9\times\frac{H}{8}\times\frac{W}{8}}=(z_i,z_m,z_s)$, and feed them to the U-Net structure in the denoising process. We only feed the input grayscale latent feature $z_i$ into the ControlNet component. The objective function then becomes: 
\begin{equation}
    \mathcal{L} = \mathbb{E}_{
        \mathcal{E}(\mathbf{x}), 
        y, 
        z_i
        \epsilon \sim \mathcal{N}(0, 1), 
        t
    } 
    \Big[\|\epsilon - \epsilon_\theta(\tilde{z}_t, t, y, z_i)\|^2_2 \Big].
\end{equation}

\noindent
\textbf{Exemplar Control.}
For exemplar-based image colorization, we add a CLIP image encoder to encode exemplars into latent features and feed them into the cross-attention blocks. During training, all the other parts of the model are fixed and only the image encoder is fine-tuned. Since there is no paired data for exemplar-based colorization, inspired by Zhang~\etal \cite{zhang2019deep}, we employ \textbf{contextual loss} to constrain color distribution of the generated results similar to the exemplars'. 
\begin{equation}
\begin{aligned}
{d}^l(i,j)&=cos(\phi^{l}_{I_e}(i),\phi^{l}_{I_g}(j)),\\
A^l(i,j)&=\mathop{softmax}\limits_{j}(1-\tilde{d}^l(i,j)/h),\\ 
L_{context}&=\sum_{l\in [3,5]} w_l[-log(\frac{1}{N_l}\sum_i \mathop{max}\limits_j (A^l(i,j))],
\end{aligned}
\end{equation}
where $\phi^l$ are the $l^{th}$ layers of the pretrained VGG19 model~\cite{simonyan2014very}, $I_e$ is the exemplar image and $I_g$ is the predicted images at diffusion time step $t$, formulated in Eq.~\eqref{eq:exemplar}. $\tilde{d}^l(i,j)$ is the normalized cosine similarity $d^l(i,j)$ of pairwise feature points. $A^l(i,j)$ denotes the pairwise affinities between features from $l^{th}$ layer. The parameter $h$ is set to $0.01$ and $w_l$ are set to $8,4,2$ for $l=5,4,3$. 

Besides, we introduce a \textbf{grayscale loss} to constrain the content of the generated result similar to the input image.
\begin{equation}
   L_{gray}=||\frac{\sum_{\{R,G,B\}}{I_{i}}}{3}-\frac{\sum_{\{R,G,B\}}{I_{g}}}{3}||_2,
\end{equation}
 where $I_i$ is the input image regenerated by the autoencoder and $I_g$ is the decoded generated result in each diffusion timestep $t$, which can be formulated as:
 \begin{equation}
 \begin{aligned}
   I_i&=\Phi_d(\Phi_e(I_{in})),\\
   I_g&=\Phi_d(\frac{X_t-\sqrt{1-\bar{\alpha}_t}\epsilon_t}{\sqrt{\bar{\alpha}_t}}),
   \label{eq:exemplar}
\end{aligned}
\end{equation}
 where $I_{in}$ is the original ground truth image, $\epsilon_t$ is the predicted noise at time step $t$, $X_t$ is the latent code of the image at $t$ in the diffusion forward process and $\Phi_e$ and $\Phi_d$ are the encoder and decoder of the fixed autoencoder. 
\textbf{The final loss function} for training exemplar-based colorization can be formulated as:
 \begin{equation}
L_{exemplar}=L_{context}+w_e*L_{gray}.
\end{equation}
We set $w_e$ to $1000$ empirically.

\vspace{-2mm}
\subsubsection{Color Overflow and Incorrect Color Handling}
\vspace{-1mm}
\label{sec:color_overflow}
\noindent\textbf{Content-guided Deformable Autoencoder.} To handle large color overflow and incorrect color regions, we introduce content-guided deformable convolution layers into the autoencoder's decoder, as shown in Fig.~\ref{fig:img_pipeline} (right).

Specifically, we add a deformable convolution layer after each of the first three convolution blocks in the decoder of the autoencoder. The input grayscale image $I_i\in\mathbb{R}^{3\times H\times W}$ is first encoded into the latent space as $z^{4\times \frac{H}{8}\times \frac{W}{8}}$ using the fixed original autoencoder's encoder and then fed into the deformable convolution as guidance. During training, we only train these three additional deformable convolution layers. The deformed color images are generated by our main model randomly using different types of conditional colorizations. The loss function is $perceptual\ loss$ for the first $500$ training steps to constrain the reconstruction and $perceptual\ loss + 0.025\times discriminator\ loss$ for the following steps.

During inference, we replace the original autoencoder's decoder with the learned deformable-enable decoder and feed the input grayscale image as guidance to it in the same way as training. In our interface, users have the option to choose whether to employ the learned decoder. The learned decoder tends to align colors within the same area, toward similarity. However, users may opt against its use to retain greater flexibility.  

\noindent\textbf{Streamlined Self-Attention Guidance.} Inspired by Hong \etal~\cite{hong2023improving} for improving sample quality of generative models, we introduce training-free guidance in the inference process to handle small color overflow, as shown in Fig.~\ref{fig:img_pipeline} (bottom right). 
This refined self-attention guidance is employed during inference to blur and re-predict small overflow areas by referencing the surrounding color distribution.

In particular, we modify the predicted noise $\epsilon_t$ using the following steps. The attention map $A_t$ is the similarity map calculated by the query $Q$ and the key $K$ after softmax in each attention block. We follow \cite{hong2023improving} to get the degraded predicted $\hat{X_0}'$ and the attention mask $M_t$, which masks out the majority of the image except for salient parts.

To better retain the original color distribution, different from Hong \etal~\cite{hong2023improving}, we preserve the degraded $\hat{X_0'}$ without additional noise to retain more unmasked color information. 
Specifically, we obtain $X_t'$ from Eq.~\eqref{eq:xt} and the modified $\hat{\epsilon_t}'$ from Eq.~\eqref{eq:et}. Then for $t\in [T, t_s]$, we use $\hat{\epsilon_t}'$ from Eq.~\eqref{eq:et} to replace $\hat{\epsilon_t}$ and output the new predicted $X_{t-1}$. Here the number of the total denoising steps $T$ is $1000$ and we set $t_s$ to $600$ empirically. 
\vspace{-1.7mm}
\begin{equation}
    X_t'\leftarrow(1-M_t)\odot X_t'+ M_t\odot\hat{X_0'},
    \label{eq:xt}
\end{equation}
\begin{equation}
    \hat{\epsilon_t}'\leftarrow model(X_t'),
\end{equation}
\begin{equation}
    \hat{\epsilon_t}'\leftarrow\hat{\epsilon_t}+s*(\hat{\epsilon_t}-\hat{\epsilon_t}'),
    \label{eq:et}
\end{equation}
where $\odot$ represents element-wise multiplication, $model$ represents our diffusion model and $s$ denotes the scale of the guidance. Based on experiments, we set $s$ to $0.05$. We offer an option to change $s$ in our interface. Further discussion on the impact of $s$ is provided in our suppl.. 

\vspace{-2mm}
\section{Experiments}
\label{sec_experiments}

\newcommand{\best}[1]{\textcolor{red}{\textbf{#1}}}
\newcommand{\second}[1]{\textcolor{blue}{\textbf{#1}}}

\subsection{Implementation}
\label{sec: imple}
\noindent
\textbf{Training.}
We implement our method with PyTorch and train it on four NVIDIA RTX 3090Ti GPUs. 
We use AdamW optimizer with $\beta_1=0.9$ and $\beta_2=0.999$.
The learning rate is set to $10^{-5}$. 
We train our model for $80K$ global steps with a batch size of 64.
The first $15K$ training is based on the Stable Diffusion v1.5 and the later $65K$ global steps add the stroke branch as shown in Fig.~\ref{fig:img_pipeline} for further finetuning.
Following Stable Diffusion~\cite{rombach2022high}, we resize the input images to $512\times512$ and convert them to $Lab$ color space, and use the $L$ channel as the input condition.
For exemplar-based colorization, we train the image encoder for $100K$ global steps.  
For the deformable convolution layers training, we train it for $9K$ global steps.

To make the model more sensitive to color, we collect a dictionary of 235 commonly used English words that describe colors for filtering. 
During the training process, we generate all the captions automatically using BLIP~\cite{li2022blip}. If a generated caption does not describe a black-and-white image (\ie, does not include phrases like ``a black and white photo of $\cdots$''), and contains color-related words, we reserve all these captions as input. 
For the remaining captions, we randomly make 60\% of them to be empty, which are used for training the unconditional generation.

To simulate both the spontaneous strokes of users and the presence of more hints in exemplars, we first utilize  Simple Linear Iterative Clustering (SLIC) \cite{achanta2012slic} to obtain a superpixel representation of the image. Then for each image, we randomly crop $1$ to $100$ areas that are rectangular regions of varying sizes from $5\times5$ to $50\times50$ pixels. There is a 20\% probability that these regions correspond to local regions from the ground truth, while the remaining 80\% are local regions from the superpixel representation. 
In addition, we apply color jittering to 20\% of the hint images and ground truth. The purpose is to guide our model to trust the hint points rather than just output normal color distribution regardless of whether hints are given. 
At this stage, the hint images are always non-empty, while the prompt input is always empty, ensuring that the model can still generate normal-colorized images during the prompt-based conditional generation process.

\noindent
\textbf{Inference.}
We use a single NVIDIA RTX 3090Ti GPU for inference. Since the diffusion model is trained on the images of size $512\times512$ and is sensitive to the resolutions, we proportionally resize the input image  to a shorter side of 512 pixels before feeding it to our model. Then, we resize it back to its original size before replacing the $L$ channel.

\begin{figure}[!t]
    \begin{center}
    \vspace{-3mm}
    \includegraphics[width=\columnwidth]{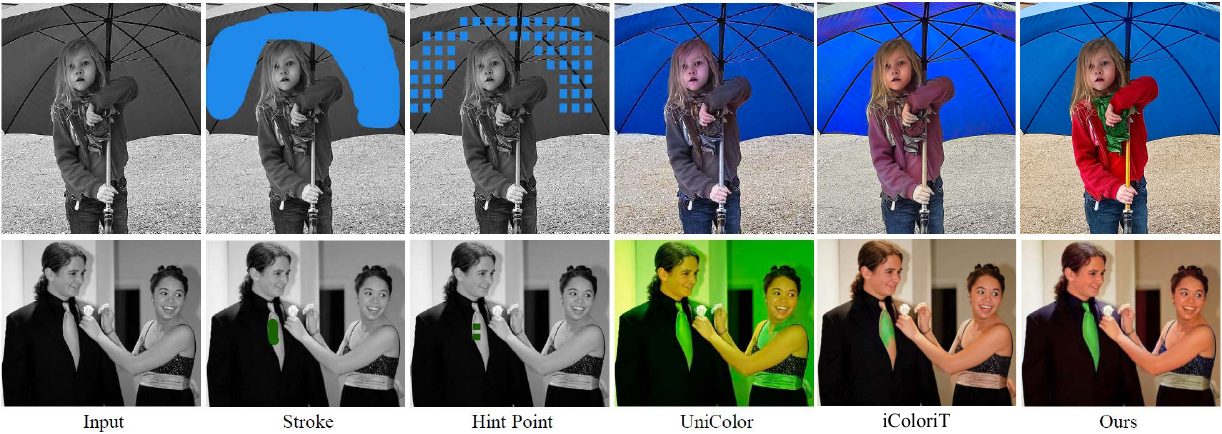}
    \end{center}
    \vspace{-6mm}
    \caption{Qualitative comparisons on stroke-based colorization.}
    \label{fig:stroke}
    \vspace{-3mm}
\end{figure}

\begin{figure}[t]
    \begin{center}
    \includegraphics[width=\linewidth]{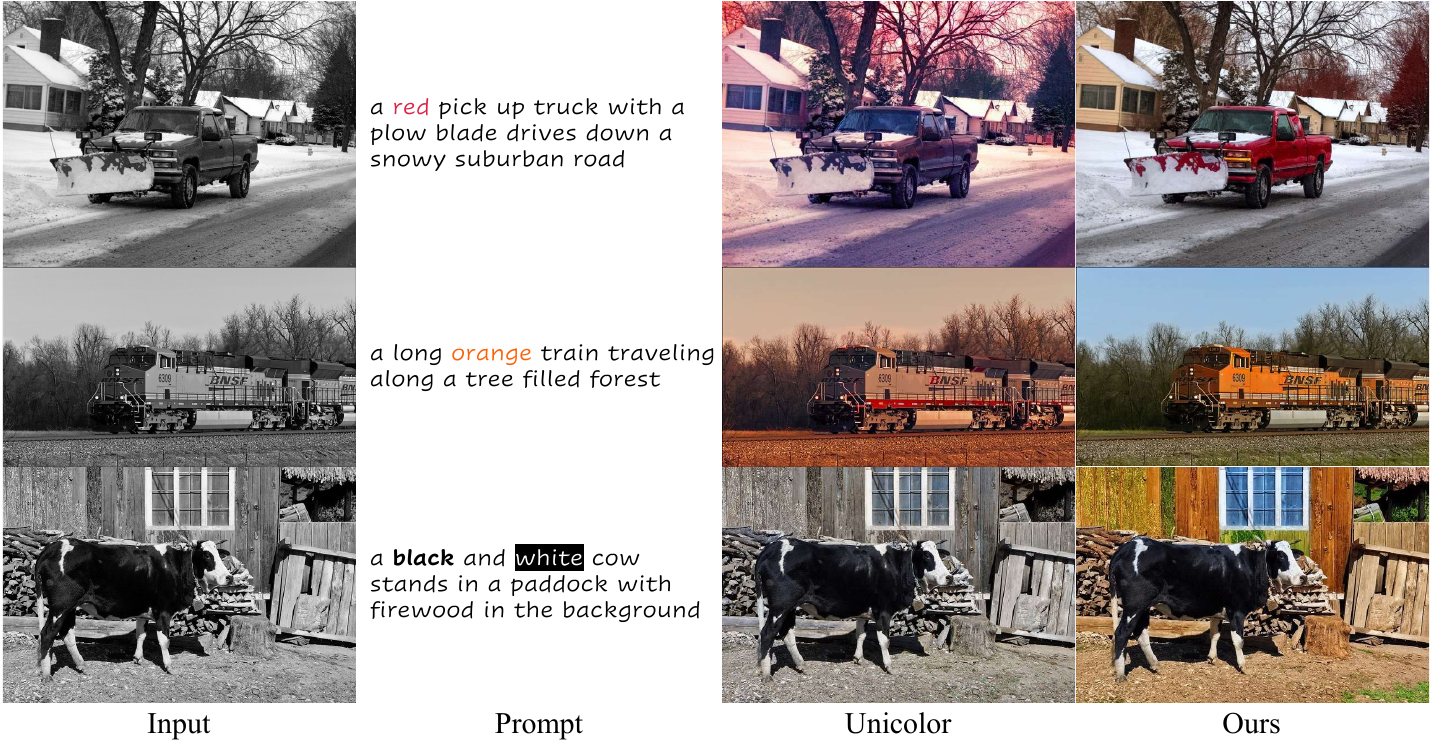}
    \end{center}
    \vspace{-6mm}
    \caption{Qualitative comparisons on prompt-based colorization. 
    }
    \label{fig:text-based}
    \vspace{-6mm}
\end{figure}

\begin{table}[!b]
\vspace{-4mm}
\caption{Quantitative results for unconditional colorization on the ImageNet val5k dataset, ImageNet ctest dataset, and COCO-Stuff dataset. ``Ours'' is purely unconditional. As for ``Ours (hint)'', we randomly give $50$-$100$ random hint points for each image (each image has more than $260K$ pixels). Results indicate the inherent color priors in our model are more colorful than ImageNet and the effectiveness of our stroke-based colorization.}
\vspace{-7mm}
  \begin{center}
  \resizebox{\columnwidth}{!}{%
  \begin{tabular}{c|cc|cc|cc}
    \toprule
     Dataset& \multicolumn{2}{c|}{ImageNet (val5k)}& \multicolumn{2}{c|}{ImageNet (ctest)} & \multicolumn{2}{c}{COCO-Stuff} \\
      Metrics& FID$\downarrow$ & Colorfulness$\uparrow$  & FID$\downarrow$ & Colorfulness$\uparrow$  & FID$\downarrow$ & Colorfulness$\uparrow$  \\
    \midrule
    CIC~\cite{zhang2016colorful} &22.0860 &37.0313& 12.7651 & 37.5761&33.3418 &37.6487  \\
    UGColor~\cite{zhang2017real} &15.1777&27.0966 & 6.5466 & 27.8122 &21.4010 &28.4487 \\
    DeOldify~\cite{antic2019deoldify} &10.5191&26.4827 & 4.2143 & 23.1538&13.4318 &28.3779 \\
    ChromaGAN~\cite{vitoria2020chromagan} & 16.4390 & 25.5862& 9.3487& 29.0895& 26.4624& 29.1411\\
    InstColor~\cite{su2020instance} &12.9455&27.5710 & 6.7803 & 28.1923&12.6844 &29.2302  \\
    ToVivid~\cite{wu2021towards} &5.8019 &37.3376 & 2.6775 & 37.8425& 8.5452& 38.8155\\
    BigColor~\cite{kim2022bigcolor} & 7.7677 & 42.5364&  2.8583 & 44.4135&10.0362 &43.4104  \\
    DISCO~\cite{xia2022disentangled} &10.2895&40.9533 & 5.7196 & 37.4613&13.3850 &39.1969 \\
    DDColor~\cite{kang2023ddcolor} &5.5726 &42.8370 &2.6294 &42.9575 &7.2718 &42.2919 \\
    ColorFormer~\cite{ji2022colorformer} &6.3831 &40.5631  &2.9816 &41.2833 &8.5623 &41.0248\\
    UniColor~\cite{huang2022unicolor} &9.6292 &36.3415 &4.9140 &37.1726 &8.0509 &36.7442\\
    \hline
    Ours &8.8749 &\textbf{47.1680} &4.2915 &\textbf{44.9256} &10.2651 &\textbf{47.0501}  \\
    Ours (hint) &\textbf{4.6657} &32.3621 &\textbf{2.0990}&33.5493  &\textbf{6.8830} &32.0113\\
    \bottomrule
  \end{tabular}
  }
  \vspace{-2mm}
  \end{center}

  \label{tab:uncond}
  \vspace{-2mm}
\end{table}

\begin{figure}[t]
    \begin{center}
    \vspace{-3mm}
    \includegraphics[width=\linewidth]{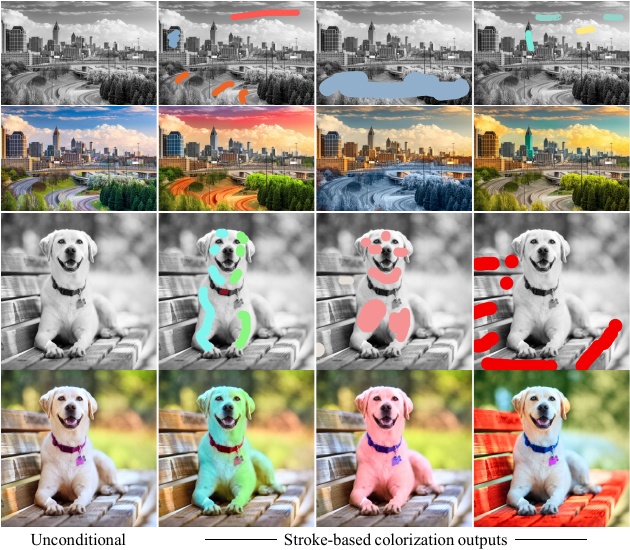}
    \end{center}
    \vspace{-6mm}
    \caption{Visual examples for our stroke-based colorization.}
    \label{fig:stroke-based}
    \vspace{-2mm}
\end{figure}
\begin{figure}[t]
    \begin{center}
    \vspace{-2mm}
    \includegraphics[width=1.0\columnwidth]{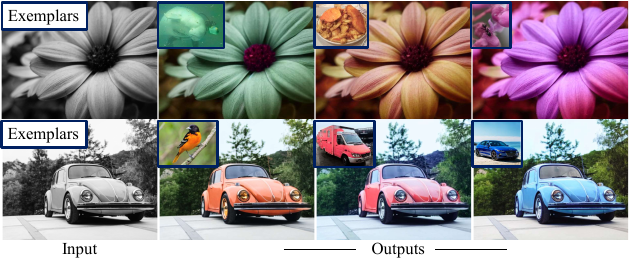}
    \end{center}
    \vspace{-6mm}
    \caption{Visual examples for our exemplar-based colorization.}
    \label{fig:exemplar-based}
    \vspace{-6mm}
\end{figure}

\subsection{Dataset}
\noindent
\textbf{Training Data.}
For training of unconditional, prompt-based, and stroke-based colorization, we use the ImageNet training dataset and filter out relatively gray images. Specifically, we calculate the mean of the variance of the difference between every two channels of the RGB images, denoted as $E(Var(C_i,C_j))$, and filter out the images whose $E(Var(C_i,C_j))$ is lower than a threshold. We set the threshold to $12$ empirically. 

For exemplar-based colorization training, we use the same ImageNet data as the inputs. Because we aim to colorize target images based on semantically similar objects in the exemplars, we utilize CLIP~\cite{radford2021learning} for corresponding exemplar retrieval. Specifically, we first use the pre-trained CLIP image encoder to get the features of each image in the dataset, then for each input image, we traverse through the features of all the other images in the same class as itself, calculate the cosine similarity between the feature of the given picture and other pictures, and then select the image with the highest similarity as the exemplar for that image.

For content-guided deformable convolution decoder training, we randomly select $100K$ images from ImageNet training dataset and use 30\% exemplar-based colorization and 70\% other types of our colorization methods to generate the images as the deformed color images to train.

\noindent
\textbf{Evaluation Data.}
For unconditional image colorization, following~\cite{kang2023ddcolor}, we perform evaluations on the whole COCO-stuff validation set~\cite{caesar2018coco} and the first 5k images in ImageNet validation dataset, \ie, val5k, and following~\cite{larsson2016learning}, we also perform evaluations on the ImageNet ctest dataset (10k images), another subset of the ImageNet validation set.
For prompt-based image colorization, following Weng \etal~\cite{weng2022code}, we use the COCO-Stuff validation set~\cite{caesar2018coco} and keep only images with adjectives in captions (2.4k images). 

\begin{table}[b]
\vspace{-5mm}
\caption{Quantitative results for prompt-based colorization on COCO-Stuff~\cite{caesar2018coco} dataset.}
\vspace{-7mm}
  \begin{center}
  \resizebox{0.8\columnwidth}{!}{%

  \begin{tabular}{c|ccc}
    \toprule
      & FID$\downarrow$ & Colorfulness$\uparrow$ & CLIP score$\uparrow$\\
    \midrule
    UniColor~\cite{huang2022unicolor} & 16.6911 & 45.7293 & 0.7649\\
    Ours & \textbf{15.2651} & \textbf{65.5266} & \textbf{0.7871}\\
    \bottomrule
  \end{tabular}

  }
  \end{center}
   
  \label{tab:text-based}
  \vspace{-5mm}
\end{table}

\begin{figure*}[t]
\vspace{-7mm}
    \begin{center}
    \includegraphics[width=1.0\linewidth]
    {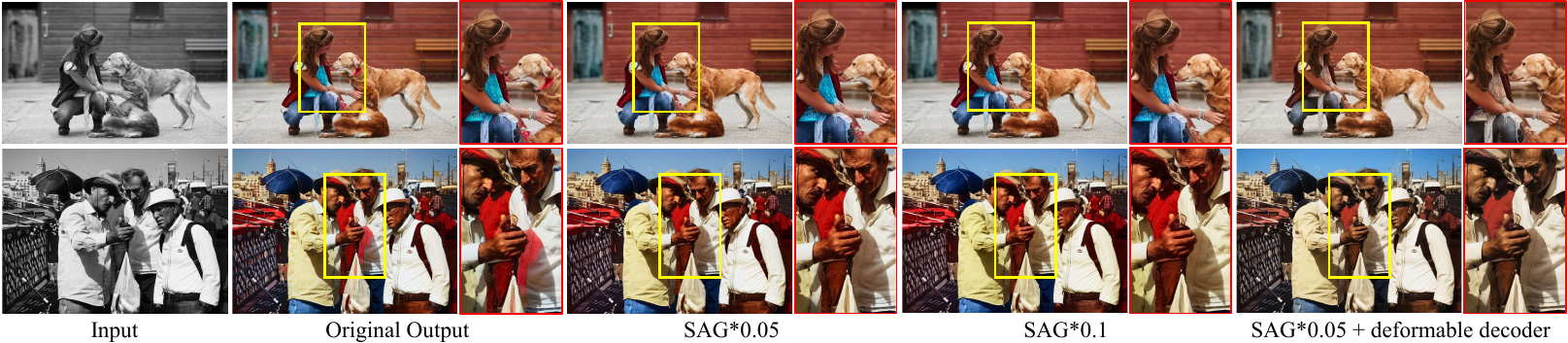}
    \end{center}
    \vspace{-6mm}
    \caption{The ablation study for content-guided deformable autoencoder decoder and streamlined self-attention guidance (SAG). The output without SAG and the deformable layers will generate incorrect colors in small objects, like the shadow, scarf, arms, hat, and so on. }
    \label{fig:sag_comparison}
    \vspace{-3mm}
\end{figure*}

\subsection{Comparisons}
For fair comparisons, we only compare our method with state-of-the-art methods that keep input and output resolutions unchanged. 
For image colorization, following Huang \etal~\cite{huang2022unicolor}, we use Fréchet Inception Distance (FID) \cite{heusel2017gans} and colorfulness \cite{hasler2003measuring} for unconditional and prompt-based colorization. 
For prompt-based methods, we also employ the CLIP score~\cite{radford2021learning}  to evaluate the proximity between colorized images and text descriptions in the CLIP latent space. 

\noindent
\textbf{Unconditional Colorization Comparison.} 
For unconditional colorization, we compare our method with two types of state-of-the-art methods: 1) six CNN-based methods: CIC \cite{zhang2016colorful}, UGColor \cite{zhang2017real}, DeOldify \cite{antic2019deoldify}, ChromaGAN \cite{vitoria2020chromagan}, InstColor \cite{su2020instance}, ToVivid \cite{wu2021towards}, and BigColor \cite{kim2022bigcolor}; 2) four Transformer-based methods: UniColor \cite{huang2022unicolor}, DISCO \cite{xia2022disentangled}, ColorFormer~\cite{ji2022colorformer}, and DDColor~\cite{kang2023ddcolor}.

For quantitative comparison, as presented in Tab.~\ref{tab:uncond}, our pure unconditional colorization achieves state-of-the-art performance in terms of colorfulness yet FID. This suggests our unconditional method generates more colorful results but the color distribution may be different from the ImageNet dataset, due to the SD model pre-trained on LAION-5B~\cite{schuhmann2022laion}, which is more colorful and high-quality than ImageNet. But with hints from ImageNet added, our method can achieve top performance in terms of FID but colorfulness becomes low. The results show that by providing some reasonable hints, our method achieves a color distribution closest to the ground truth, showing the effectiveness of our stroke-based colorization. This also indicates that ImageNet images are less colorful than the color priors in our model.

We show the visual comparison with state-of-the-art methods in Fig.~\ref{fig:uncond_compar}. 
As shown, while the compared methods tend to generate brown and greyish images, our method generates vivid and realistic colors with less color overflow.

\noindent
\textbf{Stroke-based Colorization Comparison.}
We compare visual results among our method, UniColor~\cite{huang2022unicolor}, and iColoriT~\cite{yun2023icolorit} with manually drawn strokes in Fig.~\ref{fig:stroke}. 
Our method can generate vivid and plausible colors outside the drawn object (\eg, the girl holding umbrella in the $1^{st}$ row). When the stroke is localized on a small object, our method achieves localized colorization without altering global hue (\eg, green stroke drawn on the man's tie in the $2^{nd}$ row).
Fig.~\ref{fig:stroke-based} showcases more results on stroke-based colorization.

\noindent 
\textbf{Prompt-based Colorization Comparison.} We compare our method with UniColor~\cite{huang2022unicolor}. 
L-CAD~\cite{chang2023cad} and UniColor~\cite{huang2022unicolor} are the only two methods with publicly available code for prompt-based colorization, but the output size of L-CAD~\cite{chang2023cad} is $256\times256$, so we do not compare it for fairness.
Quantitatively, as shown in Tab.~\ref{tab:text-based}, our method achieves better FID, Colorfulness, and CLIP scores than those of UniColor, suggesting that our results are closer to the ground truth images given the color-description captions and have better alignment with the given captions.
We also show the visual comparison in Fig.~\ref{fig:text-based}. 
Our method colorizes images with more vivid colors and better color consistency.

\noindent
\textbf{Exemplar-based Colorization Showcase.}
We show some visual examples in Fig.~\ref{fig:exemplar-based}. Overall, our method can colorize the inputs with a similar color distribution to exemplars while keeping the color realistic and visually pleasant.

\subsection{User Study}
\begin{figure}[t]
    \begin{center}
    \vspace{-1mm}
    \setlength{\leftskip}{-10pt}
    \includegraphics[width=1.05\linewidth]{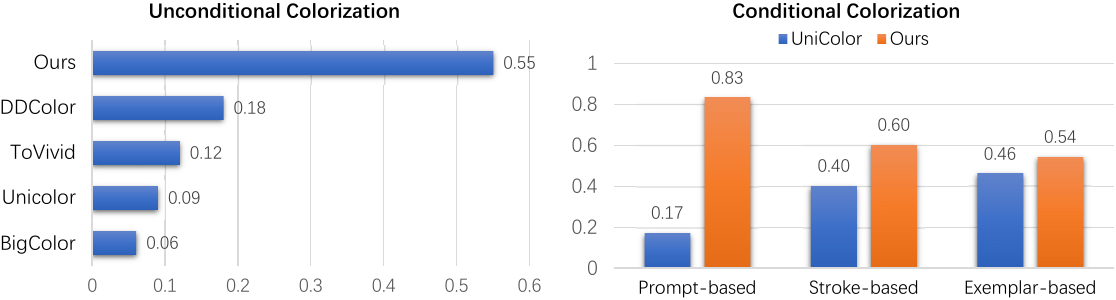}
    \end{center}
    \vspace{-6mm}
    \caption{Quantitative results of our user study.}
    \label{fig:user_study}
    \vspace{-5mm}
\end{figure}

Since colorization algorithms are mainly applied in the field of creative arts and old photo restoration, human satisfaction is always a more reasonable indicator of the algorithm's performance. 
We conduct a user study to further show the superior performance of our method. 
Our study is divided into four sections corresponding to unconditional, prompt-based, stroke-based, and exemplar-based colorization, respectively. 
Each section consists of 10 questions. 
In each question, participants are asked to select the \textit{best} image among several compared methods. 
We define \textit{best} based on the following metrics: perceptual realism (less color overflow), color richness, aesthetic sense, and faithful reconstruction given the input condition (if available). 
We select DDColor \cite{kang2023ddcolor}, ToVivid \cite{wu2021towards}, BigColor \cite{kim2022bigcolor}, and UniColor \cite{huang2022unicolor} for unconditional colorization, and UniColor \cite{huang2022unicolor} for three condition-based colorization.
We obtain 33 valid responses for unconditional colorization and 41 valid responses for conditional colorization. The statistic is shown in Fig.~\ref{fig:user_study}. 
Our method outperforms all the other methods.

\subsection{Ablation Study}
We carry out ablation studies to further analyze our designs.

\noindent\textbf{The impact of stroke numbers.} 
We conduct a quantitative study on the COCO-stuff dataset with randomly selected $16\times16$ cell regions. The color of the hints is derived from the mean of the superpixels in the region. 
As shown in Tab.~\ref{tab:num_stroke}, the more strokes are (as reflected by the number of cells), the closer the colorized image is to ground truth as reflected by the FID, PSNR~\cite{huynh2008scope}, SSIM~\cite{wang2004image}, and LPIPS~\cite{zhang2018unreasonable} values. There is also a trade-off between image colorfulness and faithful color reconstruction given stroke colors, as more stroked regions mean less space left to exploit the rich color priors in our model.

\noindent\textbf{The effectiveness of content-guided deformable convolution layer and streamlined self-attention guidance.}
As shown in Fig.~\ref{fig:sag_comparison}, in the original results, objects are miscolored when segmented, shadows are also incorrectly colored, and there is slight color overflow. For example, in the first row, the lower part of the woman's scarf is white, but the upper part is colored blue; the dog's neck, legs, and the person's arms have small areas of miscoloring or overflow. 
In the second row, shadows are colored red, there's bleeding on the bag, and inconsistency on the hat, where the brim is red while the top is brown. After adding streamlined self-attention guidance, minor bleeding or miscoloring in small areas was fixed. Moreover, using the deformable decoder unifies the colors of the same object and corrects large-region miscoloring. 

\label{sec:expriment}

\begin{table}
\vspace{-1mm}
\caption{The impact of the number of strokes.}
\vspace{-2mm}
  \resizebox{\columnwidth}{!}{%
  \begin{tabular}{c|ccccc}
    \toprule
    number of cropped cells & FID$\downarrow$ & Colorfulness$\uparrow$ & PSNR$\uparrow$ & SSIM$\uparrow$ & LPIPS$\downarrow$\\
    \midrule
    0 - 20 & 8.8614 &\textbf{36.3481}  &24.4910 &0.9918 &0.1305\\
    20-50 &7.8182 & 32.9553 & 26.6695 &0.9924 & 0.1034 \\
    50 - 100 & \textbf{6.8830} & 32.0113 & \textbf{28.2699} & \textbf{0.9926} & \textbf{0.0872} \\
    \bottomrule
  \end{tabular}   
  }

  \label{tab:num_stroke}
  \vspace{-4mm}
\end{table}

\section{Conclusion}
\label{sec_conclusion}
In this paper, we successfully achieve highly controllable multi-modal image colorization. Our method is built on a pre-trained SD model with new designs such as stroke control and exemplar control, which offer a diverse range of colors and high user interactivity.  With our content-guided deformable convolution autoencoder decoder and streamlined self-attention guidance, we effectively mitigate color overflow and miscoloring. Our study is an early attempt to employ SD model to handle multi-control colorization and deal with miscoloring. We believe our work can provide significant insights into controllable image colorization.

\clearpage
{
    \small
    \bibliographystyle{ieeenat_fullname}
    \bibliography{main}
}

\clearpage
\maketitlesupplementary
In this supplementary material, we present various applications of our method (Sec. \ref{sec_suppl:applications}), more ablation studies (Sec. \ref{sec_suppl:ablation}), more comparisons (Sec.~\ref{sec_suppl:comparison}), more results (Sec.~\ref{sec_suppl:results}), the reproducibility explanation (Sec.~\ref{sec_suppl:reproduce}), and the limitations of our approach (Sec.~\ref{sec_suppl:limitations}). 
In addition, a video demo is provided to showcase the interactive interface and the effectiveness and high controllability of our method at \url{https://youtu.be/tSCwA-srl8Q}.
\section{Applications}
\label{sec_suppl:applications}

\subsection{Interactive Interface}
\label{sec:interface}
\begin{figure*}[hb]
    \begin{center}
    \includegraphics[width=1.0\linewidth]{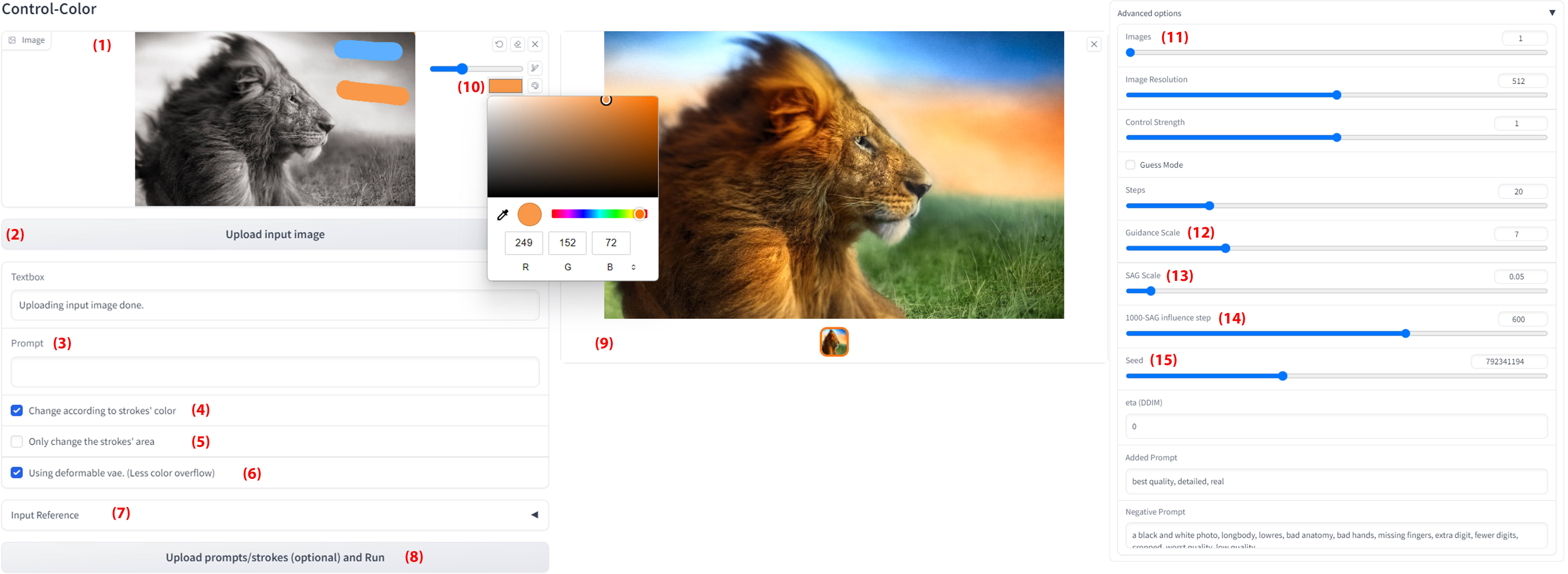}
    \end{center}
    \vspace{-4mm}
    \caption{The interactive interface of our CtrlColor. Check out Sec.~\ref{sec:interface} for more details.}
    \label{fig:interface}
\end{figure*}

\begin{figure*}[t]
    \centering
    \includegraphics[width=\textwidth]{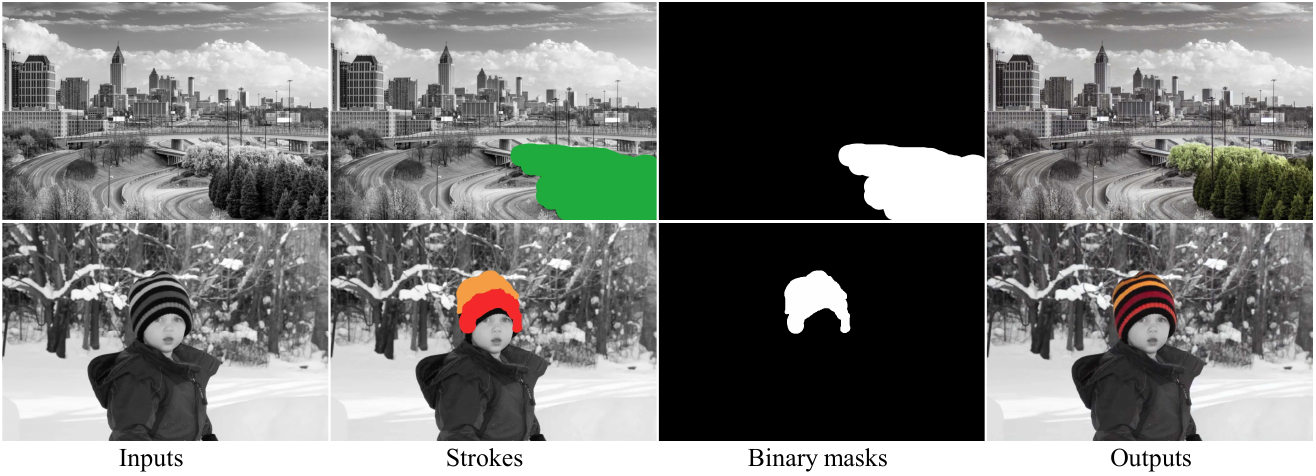}
    \vspace{-5mm}
    \caption{Showcase of our conditional region colorization. For conditional colorization, we input both the color of the strokes and the masks into our model.}
    \label{fig:region1}
    \vspace{-2mm}
\end{figure*}

As shown in Fig.~\ref{fig:interface}, we implement the interactive interface of our method using Gradio. 
The user first provides the image that needs to be colorized at (1), then just clicks the button (2)  before any other steps. 
The user can flexibly give text prompts at (3), draw strokes using (10), or give exemplar at (7). 
By choosing option (4), the image can be colorized according to the strokes' color, otherwise, it will be colorized unconditionally. 
By choosing option (5), only the color of the strokes covered areas is changed, otherwise, the whole image is re-colorized. 
To be noted, by choosing both options (4) and (5), our method colorizes the strokes covered areas according to the color of the strokes while colorizing the covered areas unconditionally if one only chooses option (5). 
By choosing option (6), the original autoencoder will be replaced by a content-guided deformable autoencoder and the latent outputs will be decoded by the learned content-guided deformable autoencoder's decoder, otherwise, the latent denoised outputs will be decoded by the original autoencoder's decoder.
Option (4) is True, option (5) is False, and option (6) is True by default. 

\begin{figure*}[t]
    \centering
    \includegraphics[width=\textwidth]{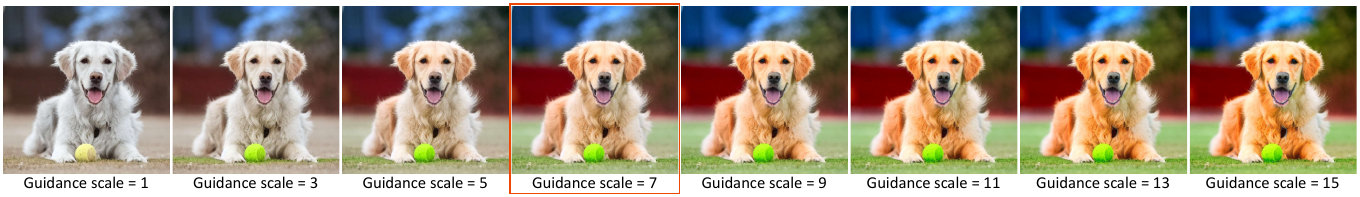}%
    \vspace{-2mm}
    \caption{The impact of the guidance scale. The higher the guidance scale, the more saturated the outputs. In our experiments, we set this scale to $7$. We offer the option in our interface to let  users adjust the saturation of the outputs based on their taste.}
    \label{fig:saturation}
    \vspace{-3mm}
\end{figure*}

The screenshot of the advanced options is shown at the right of Fig.~\ref{fig:interface}. 
Slider (11) is for changing the quantity of outputs generated each time.
The guidance scale of the conditions and the global seed can be changed using the sliders (12) and (15). Increasing the guidance scale leads to higher saturation in the color of the outputs, as shown in Fig.~\ref{fig:saturation}.
The sliders (13) and (14) are designed to change the control strength of the streamlined self-attention guidance (SAG). SAG scale (13) is the guidance scale of the SAG, and the range $[0.05,0.15]$ is recommended based on our experiments. (14) is the $t_s$ of Fig. 2 (bottom right) in our main paper. We only change the $\hat{\epsilon}_t$ between time step $t\in[T,t_s]$ during inference, and the range $[400, 600]$ is recommended. 
The parameters of the advanced options in Fig.~\ref{fig:interface} are the default settings, except for the seed which is randomly initialized.

After setting all the conditions and options, the final results will be shown in (9) after clicking the button at (8).

\subsection{Multi Controls Colorization}
As shown in Fig. 1 in our main paper and Fig.~\ref{fig:recolorization} in this supplementary material, our method supports a mix of text prompts, strokes, and exemplar controls. Our method trusts the strokes most to keep the high and precise controllability because our framework structure always keeps the latent code of the masked regions the same as the input strokes.

\subsection{Recolorization}
Since our model only takes the $L$ channel as its input, it can recolorize color images in the same manner as it does with grayscale images, as shown in Fig.~\ref{fig:recolorization}.

\subsection{Local Region Colorization}
\label{sec:reg_color}
Our method can colorize only part of the input image while keeping the other areas the same as the input. This is achieved by the design of our stroked control approach, specifically, by giving the inverse binary mask as the strokes' mask and giving the unmasked area as the hints, as shown in Figs.~\ref{fig:region1} and \ref{fig:region2} and our video demo.

\subsection{Iterative Image Editing}
With the same seed, users can easily obtain stable results. 
Users can also modify the result by adding strokes to it.  
As shown in our video demo, we can iteratively edit the image by adding different conditions. To be noted, the strokes can be covered by new strokes. Moreover, the text prompts can also be changed during editing. 

\subsection{Video Colorization}
Our methods can easily be extended to video colorization using the feature-matching method, such as LightGLUE~\cite{lindenberger2023lightglue}, to propagate the first frame to the whole video. We provide some results in the video demo.

\section{More Explanations on Key Designs} 
The inherent characteristic of the diffusion model is to regenerate the content details of an image. These deformations manifest in the final coloring as incorrect color or color overflow problems. 

We use the streamlined SAG model to solve the small color overflow problem. The mask obtained from SAG can distinguish between high and low-frequency details, with high-frequency details (such as edges) being prone to deformation, leading to areas of color overflow. We utilize this mask to blur out these high-frequency details and re-predict them according to the surrounding color distribution, as shown in Fig.~\ref{fig:SAG_scale}, thereby improving the small color overflow and incorrect color areas caused by the small deformations. 

 \begin{figure}[b]
\vspace{-6mm}
\centering
\includegraphics[width=0.7\linewidth]{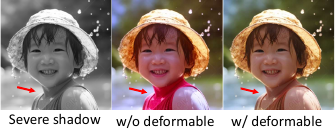}
\vspace{-2mm}
\caption{The effectiveness of the deformable layers.}
\label{fig:deformable_suppl}
\vspace{-3mm}
\end{figure}

\begin{figure*}[t]
    \centering
    \includegraphics[width=0.97\textwidth]{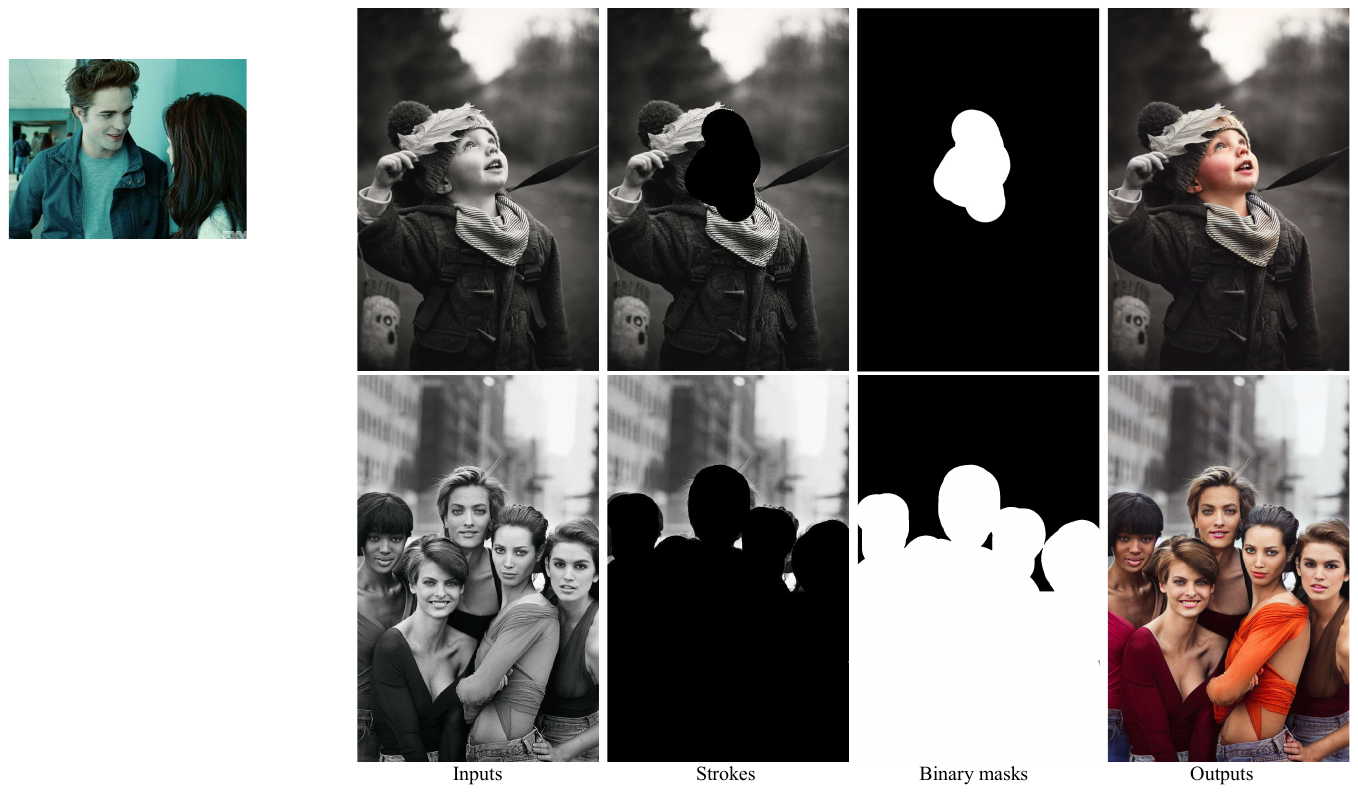}%
    \vspace{-1mm}
    \caption{Showcase of our unconditional region colorization. For unconditional region colorization, we do not input the strokes' color while only taking the masks as the conditions.}
    \label{fig:region2}
    \vspace{-4mm}
\end{figure*}
\begin{figure*}[t]
    \begin{center}
    \includegraphics[width=1.0\linewidth]
    {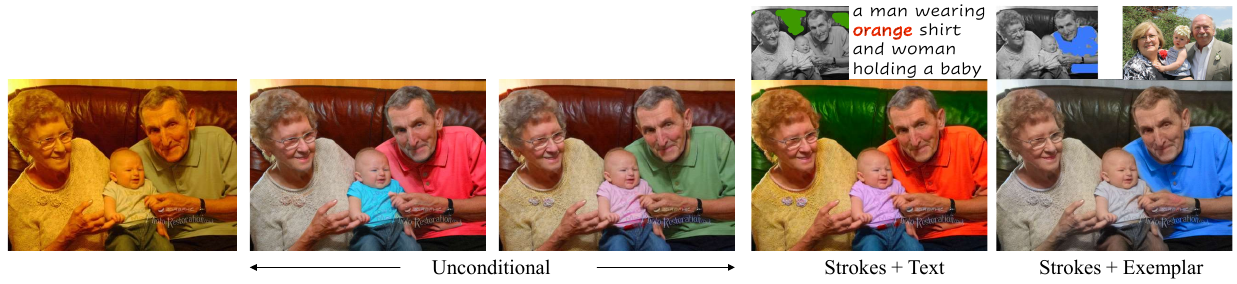}
    \end{center}
    \vspace{-5mm}
    \caption{Recolorization and multi controls colorization.}
    \label{fig:recolorization}
    \vspace{-4mm}
\end{figure*}
\begin{figure*}[t]
    \centering
    \includegraphics[width=\textwidth]{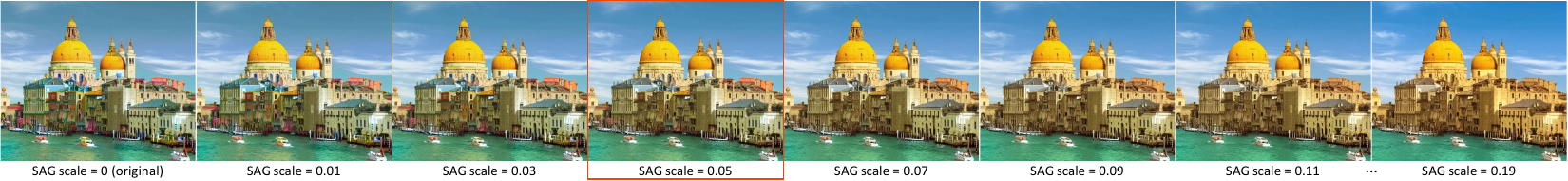}%
    \vspace{-1mm}
    \caption{The impact of the SAG scale. Here, all the $t_s$ are set to $600$. The original output (the leftmost one) has many color overflow regions on small objects. The output with a large SAG scale (\eg the rightmost one) tends to colorize large contiguous areas with the same color. Thus, we choose a scale value that effectively eliminates color bleeding without excessively blending the colors, \ie $s=0.05$.}
    \label{fig:SAG_scale}
\end{figure*}

However, the training-free approach may not solve it completely and there are more common and larger miscoloring problems, such as incorrect color caused by the mirror reflection or the object divided into two separate pieces by another object. Thus, we introduce the learnable content-guided deformable layers to correct the larger incorrect color regions. The deformable convolution layer learns an offset according to the given content, allowing the layer to perceive a larger range of features. Consequently, it corrects deformations and inaccurate color regions based on the surrounding color distribution and content. We leverage it to learn and correct inaccurate information, ensuring consistency in color with the surrounding areas. As shown in Fig.~\ref{fig:deformable_suppl}, this approach further addresses more severe incorrect color and color overflow issues.

\section{More Ablation Studies}
\label{sec_suppl:ablation}

\noindent
\textbf{Difference from the original SAG.}
After blurring unmasked areas in $\hat{X}_0$, we preserve the degraded $\hat{X}_0$ without additional noise to retain more unmasked color information, which is the key difference from the original SAG~[2]. As shown in Tab.~\ref{tab:SAG_comparison}, our refined SAG outperforms the original one in most of the metrics. This indicates our streamlined SAG can keep the original color distribution better than the original SAG.
\begin{table}
  \caption{The comparison between original and streamlined SAG on ImageNet val5k dataset.}
  \vspace{-1mm}
  \resizebox{\columnwidth}{!}{%
\begin{tabular}{c|ccccc}
    \toprule
     SAG type  & Colorfulness$\uparrow$ & PSNR$\uparrow$ & SSIM$\uparrow$ & LPIPS$\downarrow$\\
    \midrule
    original  &47.1347  &21.0038 &0.9873 &0.1946\\
    streamlined  &\textbf{47.1680}  & \textbf{21.0409} &0.9873 &0.1946 \\
    \bottomrule
\end{tabular}   
}

\label{tab:SAG_comparison}
\end{table}

\noindent
\textbf{Effectiveness of Self-Attention Guidance and Content-guided Deformable Convolution Autoencoder.} 
The quantitative comparisons in Tab.~\ref{tab:deformable} show that under the same conditions, with the self-attention guidance and the deformable convolution decoder, our model can generate color more naturally while still maintaining colorfulness, indicating fewer miscoloring problems.
\begin{table}
  \caption{The impact of the self-attention guidance and the deformable convolution layers, tested on ImageNet val5k dataset.}
  \vspace{-1mm}
  \resizebox{\columnwidth}{!}{%
  \begin{tabular}{c|cc|ccccc}
    \toprule
      &w/ SAG &w/ Deformable layers& FID$\downarrow$ & Colorfulness$\uparrow$ & PSNR$\uparrow$ & SSIM$\uparrow$ & LPIPS$\downarrow$\\
    \midrule
    \multirow{4}*{Unconditional} & &&9.7989 &59.8463&19.4400&0.9789 &0.2414\\
    &\checkmark &&11.2395 &\textbf{61.2139} &19.3985 &0.9810 &0.2332\\
    & &\checkmark &\textbf{8.3047} &28.6752 &\textbf{21.5241} &0.9870 &\textbf{0.1873}\\
    &\checkmark&\checkmark &8.8749 &47.1680 &21.0409 &\textbf{0.9873} &0.1946 \\
    \hline
    \multirow{4}*{Stroke-based}&&&5.2778&37.4117 &26.6739 &0.9887 &0.1007\\
    &\checkmark & &5.3082 &\textbf{38.6502} &27.6023 &0.9901 &0.0911\\
    & &\checkmark &5.0103 &28.6752 &27.7384 &0.9908 &0.0927\\
    &\checkmark &\checkmark &\textbf{4.6657} &32.3621 &\textbf{28.4350} &\textbf{0.9914} &\textbf{0.0876}\\
    \bottomrule
  \end{tabular}   
  }

  \label{tab:deformable}
  \vspace{-2mm}
\end{table}

\noindent
\textbf{The impact of the $s$ and the $t_s$ in our self-attention guidance.} 
The SAG scale is the $s$ in Eq.~\eqref{eq:et}. As shown in Fig.~\ref{fig:SAG_scale}, as the scale $s$ increases, the colors in larger adjacent regions tend to become more similar, which aligns with our expectations. We opted for a scale value that effectively eliminates color bleeding without excessively blending the colors, i.e. $s=0.05$.

As mentioned in Sec.~\ref{sec:color_overflow}, we only change the $\hat{\epsilon}_t$ between time step $t\in[T,t_s]$ during inference. Because the SAG process will increase the inference time and based on our experiments if $t_s\in[0,600]$, the results won't have any visually visible difference when $s=0.05$, thus, we choose $t_s=600$ to minimize the potential increase in inference time.
However, for these two parameters, especially $s$, different input images may have slightly different performances, so we provide the options in our interface.
\section{More Comparisons}
\label{sec_suppl:comparison}
As shown in Figs.~\ref{fig:comp1}, \ref{fig:comp8}, \ref{fig:comp2}, \ref{fig:comp3}, \ref{fig:comp4}, \ref{fig:comp5}, \ref{fig:comp6}, and \ref{fig:comp7}, for unconditional colorization, we provided complete visual comparisons between our method and all the other methods mentioned in Tab. 2 in our main paper.
\section{More Results}
\begin{figure*}[t]
    \centering
    \includegraphics[width=\textwidth]{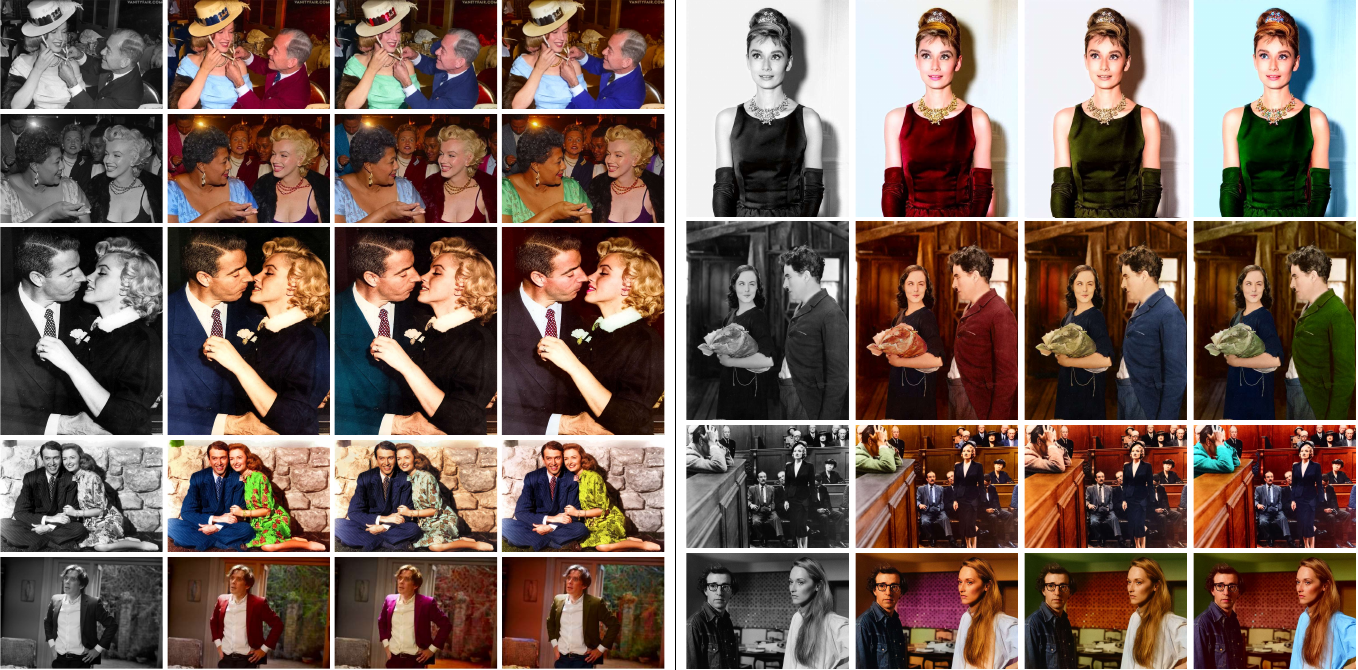}
    \vspace{-5mm}
    \caption{Showcase of our diverse image colorization on old photos and black and white films. }
    \label{fig:film_uncond}
    \vspace{-2mm}
\end{figure*}
\begin{figure*}[t]
    \centering
    \vspace{-15mm}
    \includegraphics[width=0.9\textwidth]{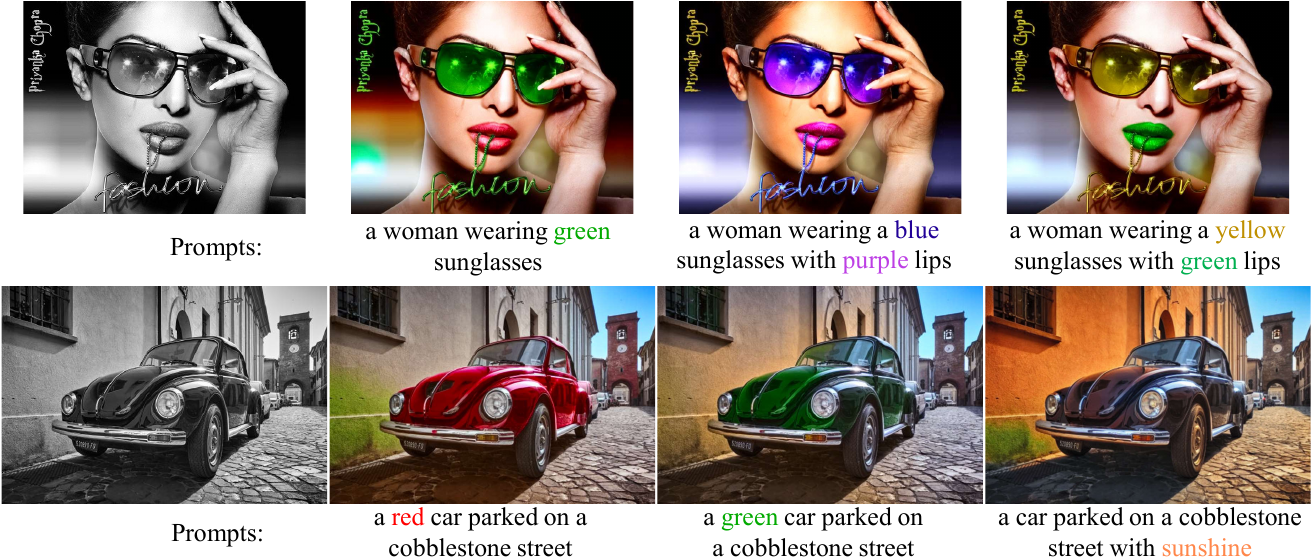}
    \caption{Visual samples of our prompt-based colorization. Our model can colorize images based on text and can even incorporate words that describe the weather.}
    \label{fig:prompts}
\end{figure*}

\label{sec_suppl:results}
As shown in Fig.~\ref{fig:film_uncond}, we provide more visual results of our diverse unconditional image colorization on old photos and black and white films.

In Fig.~\ref{fig:prompts}, we provide some visual samples of our prompt-based image colorization. 

Here we also provide more visual results to showcase the high controllability of our stroke-based colorization, as shown in Fig.~\ref{fig:stroke_ours}. Notably, \textbf{our stroke-based colorization can modify any color details precisely} if the strokes are small enough, so that, even if there is any failure case that has a small color overflow or incorrect color, users can use our stroke-based method to fix it easily. 
\section{About Reproducibility}
\label{sec_suppl:reproduce}
To ensure the reproducibility of our results, all our metrics in the document are calculated using results generated under $seed=859311133$.
\begin{figure*}[t]
\vspace{-3mm}
    \begin{center}
    \includegraphics[width=\linewidth]{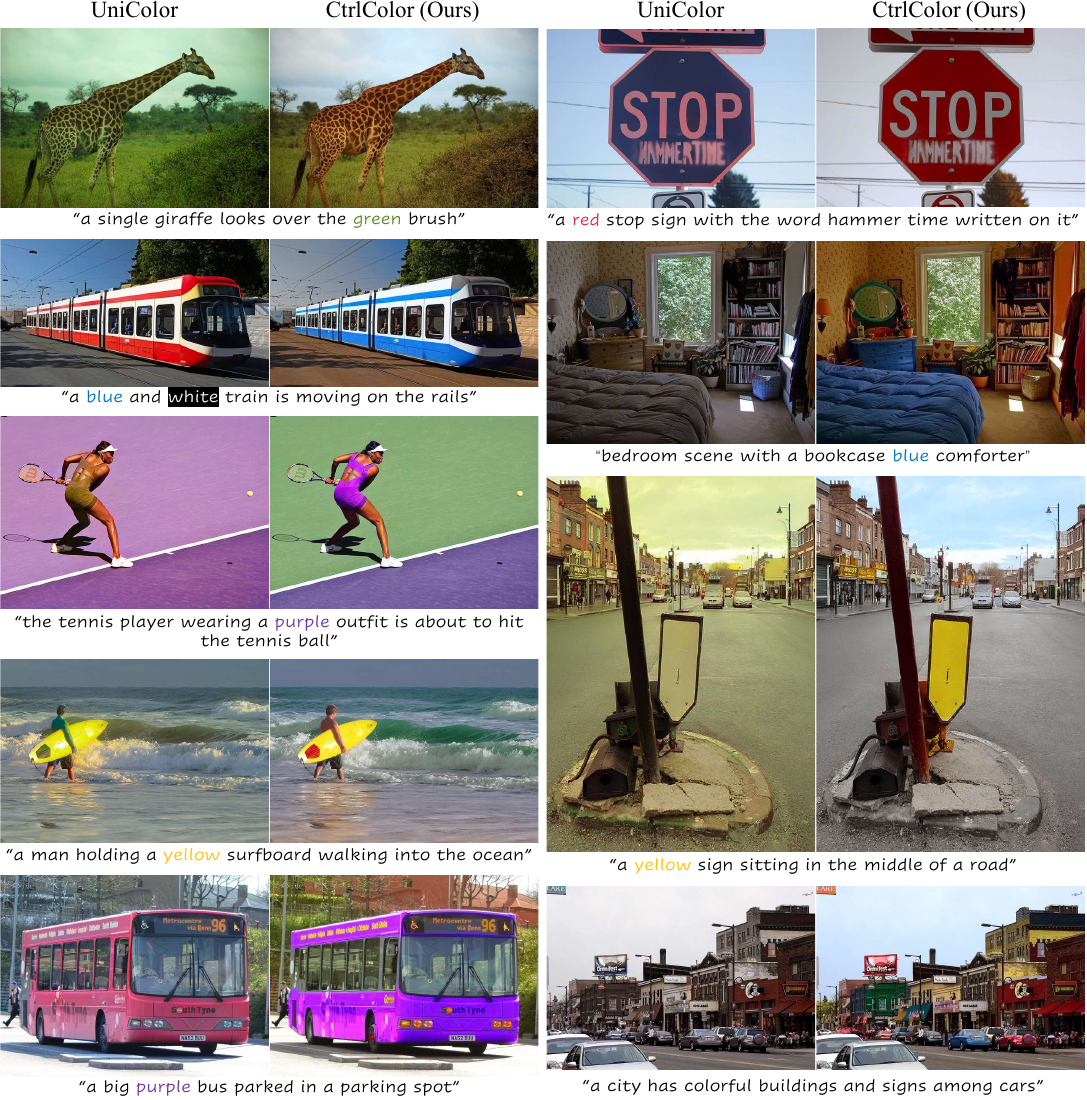}%
    \end{center}
    \vspace{-5mm}
    \caption{More visual comparisons between Unicolor~\cite{huang2022unicolor} (left) and our CtrlColor (right). Our approach can generate vivid images that closely align with the given prompts.}
    \label{fig:stroke_ours}
\end{figure*}

\section{Limitations}
\label{sec_suppl:limitations}
For region coloring,  our method may fail to generate very colorful regions if the original input image is grayscale and the region is relatively small. This is because we treat the other gray regions as the hints, so that the coloring for the specific region may be influenced by the gray hints. 
The result of the exemplar-based colorization may not be very similar to the exemplar if the exemplar's color is very complex. 
Although our method has shown promising and robust performance on image colorization, which is rarely achieved by existing methods. In some cases, our results may still have a color bleeding problem on very small objects or when the strokes are over the boundary. 
Fortunately, these issues can be solved by drawing a stroke on the bleeding area, as shown in our demo video.

\begin{figure*}[t]
\vspace{-5mm}
    \begin{center}
    \includegraphics[width=1.0\linewidth]{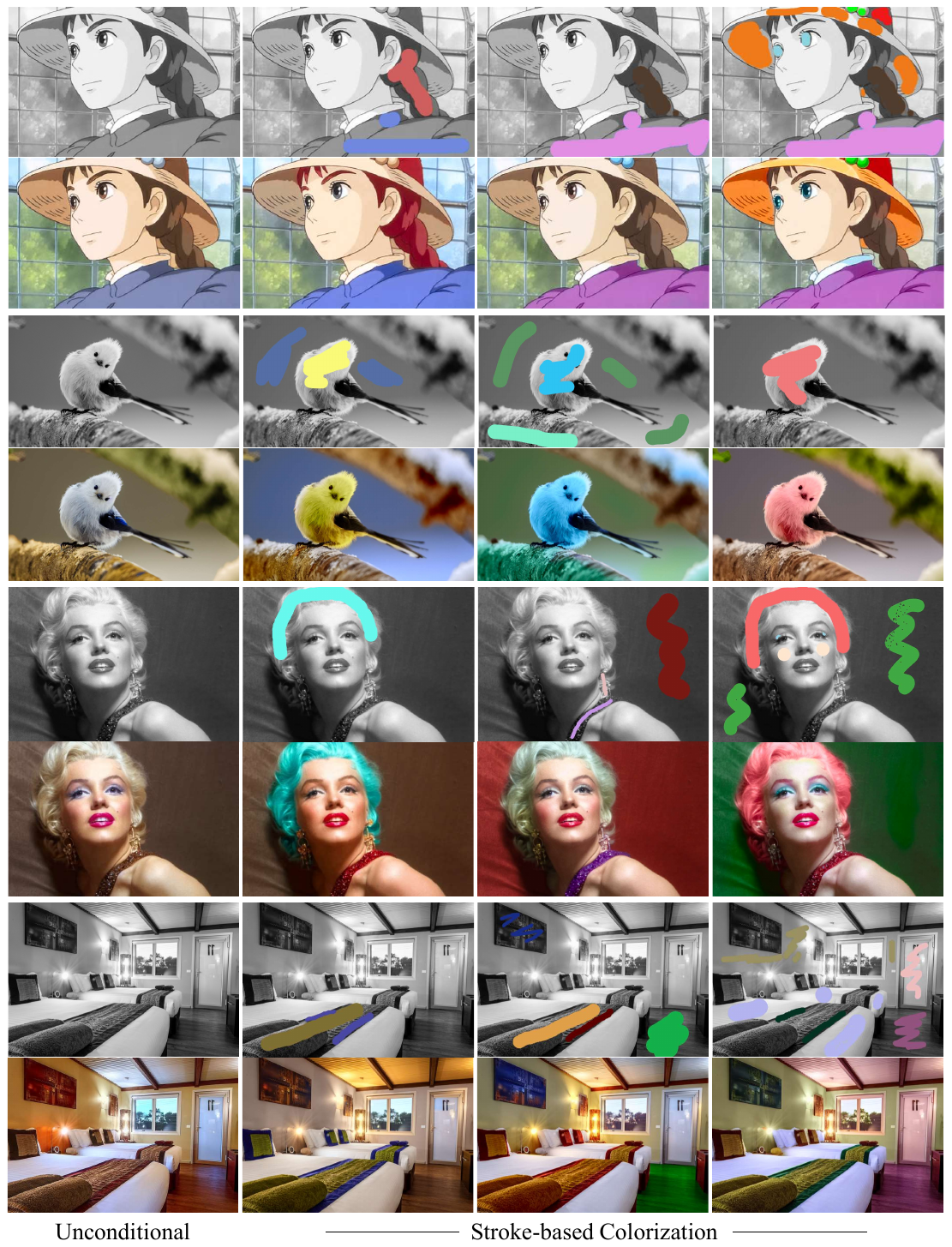}
    \end{center}
    \vspace{-4mm}
    \caption{Visual samples of our stroke-based colorization. Our approach can generate vivid images that closely align with the given strokes.}
    \label{fig:stroke_ours}
\end{figure*}

\begin{figure*}[t]
    \begin{center}
    \includegraphics[width=1.0\linewidth]{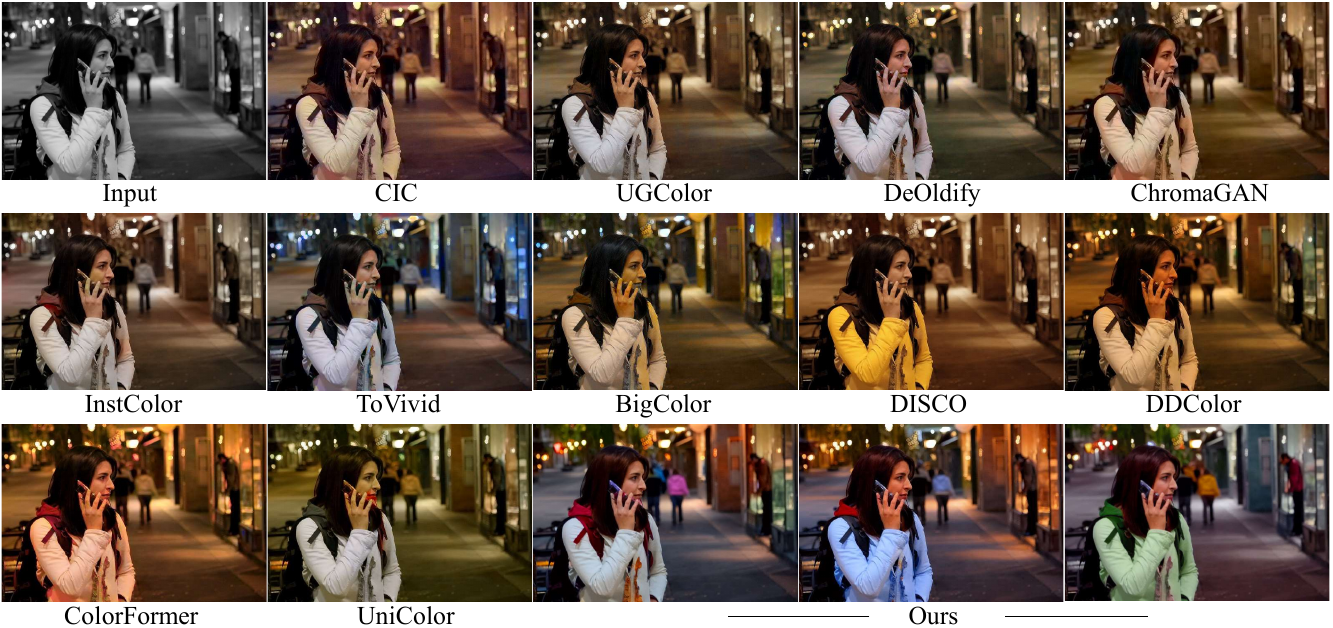}
    \end{center}
    \caption{Comparison with previous colorization methods on COCO validation set. Our CtrlColor produces images with the most color variation, separating the foreground person and the background shop windows.}
    \label{fig:comp1}
\end{figure*}

\begin{figure*}[t]
    \begin{center}
    \includegraphics[width=1.0\linewidth]{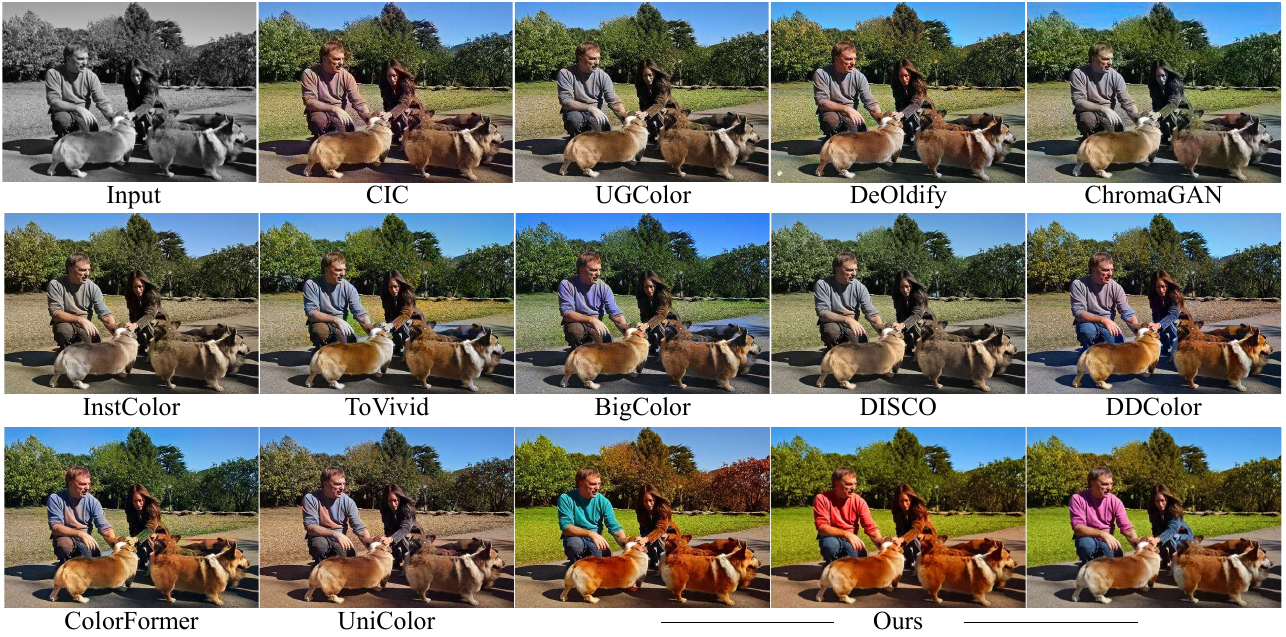}
    \end{center}
    \caption{Comparison with previous colorization methods on ImageNet validation set. Our CtrlColor is able to generate diverse and vivid colors, specially notable between the clothes wore by the two persons.}
    \label{fig:comp8}
\end{figure*}

\begin{figure*}[t]
    \begin{center}
    \includegraphics[width=1.0\linewidth]{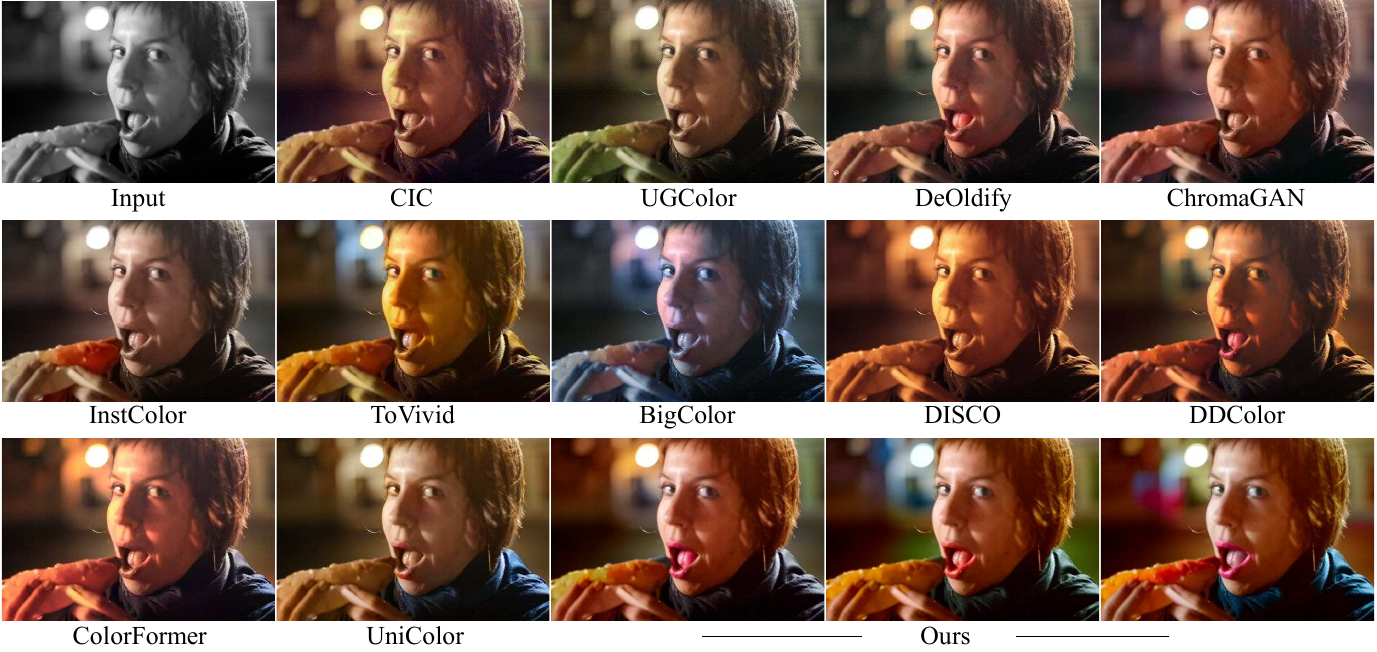}
    \end{center}
    \vspace{-3mm}
    \caption{Comparison with previous colorization methods on COCO validation set. Our CtrlColor differentiates color of tongue and lips from face, making the images more vivid and realistic.}
    \label{fig:comp2}
\end{figure*}

\begin{figure*}[t]
    \begin{center}
    \includegraphics[width=1.0\linewidth]{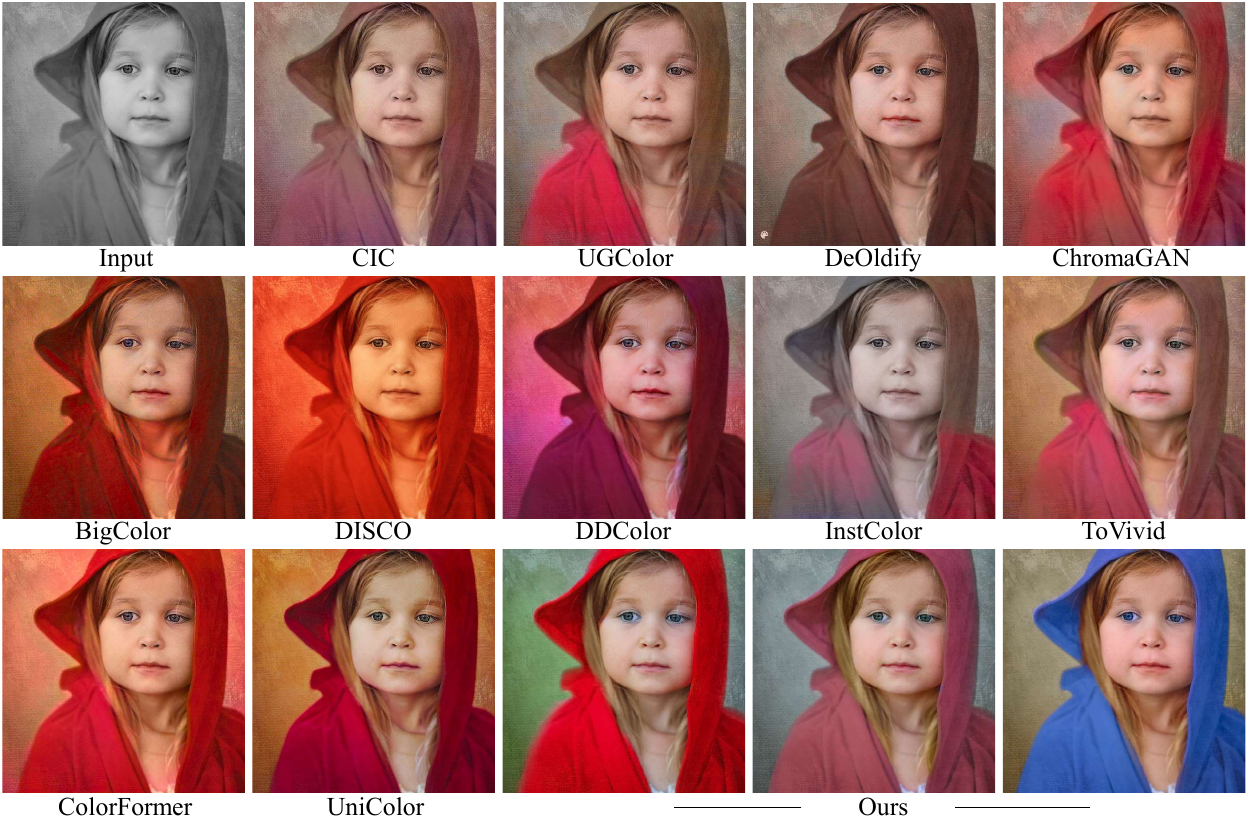}
    \end{center}
    \vspace{-3mm}
    \caption{Comparison with previous colorization methods on ImageNet validation set. Our CtrlColor produces more distinct colors and enhances the differentiation in details without color overflow.}
    \label{fig:comp3}
\end{figure*}

\begin{figure*}[t]
    \begin{center}
    \includegraphics[width=1.0\linewidth]{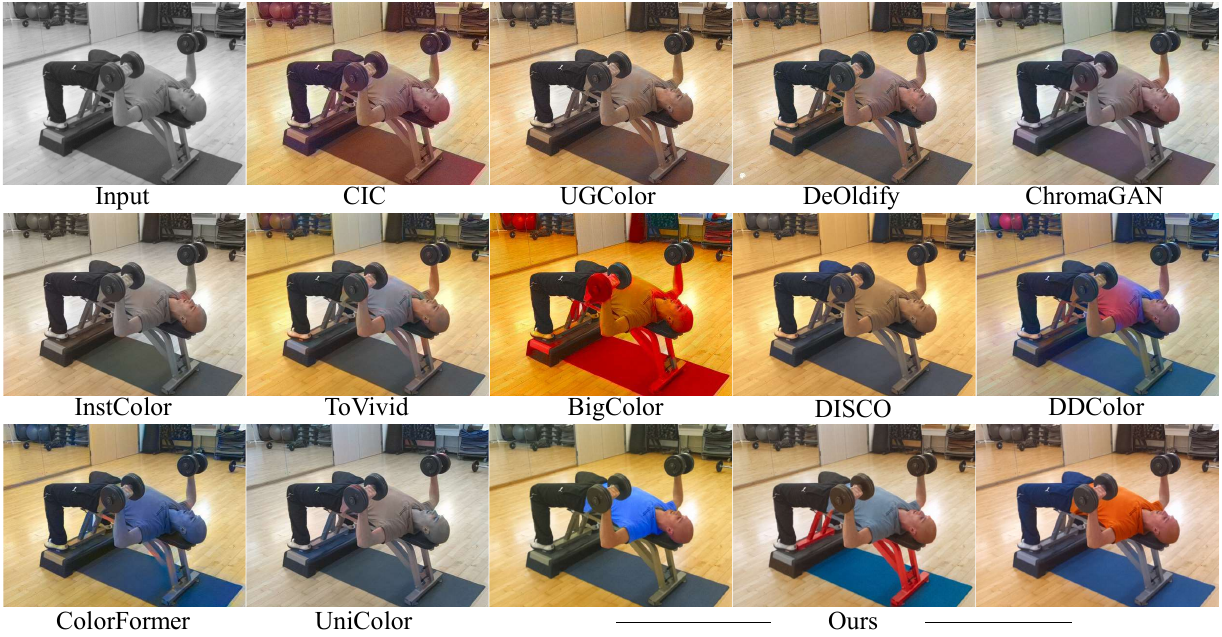}
    \end{center}
    \caption{Comparison with previous colorization methods on ImageNet validation set. Our CtrlColor generates images with distinct color boundary between skin and clothes as well as equipment and exercise mat, with no color overflow.}
    \label{fig:comp4}
\end{figure*}

\begin{figure*}[t]
    \begin{center}
    \includegraphics[width=1.0\linewidth]{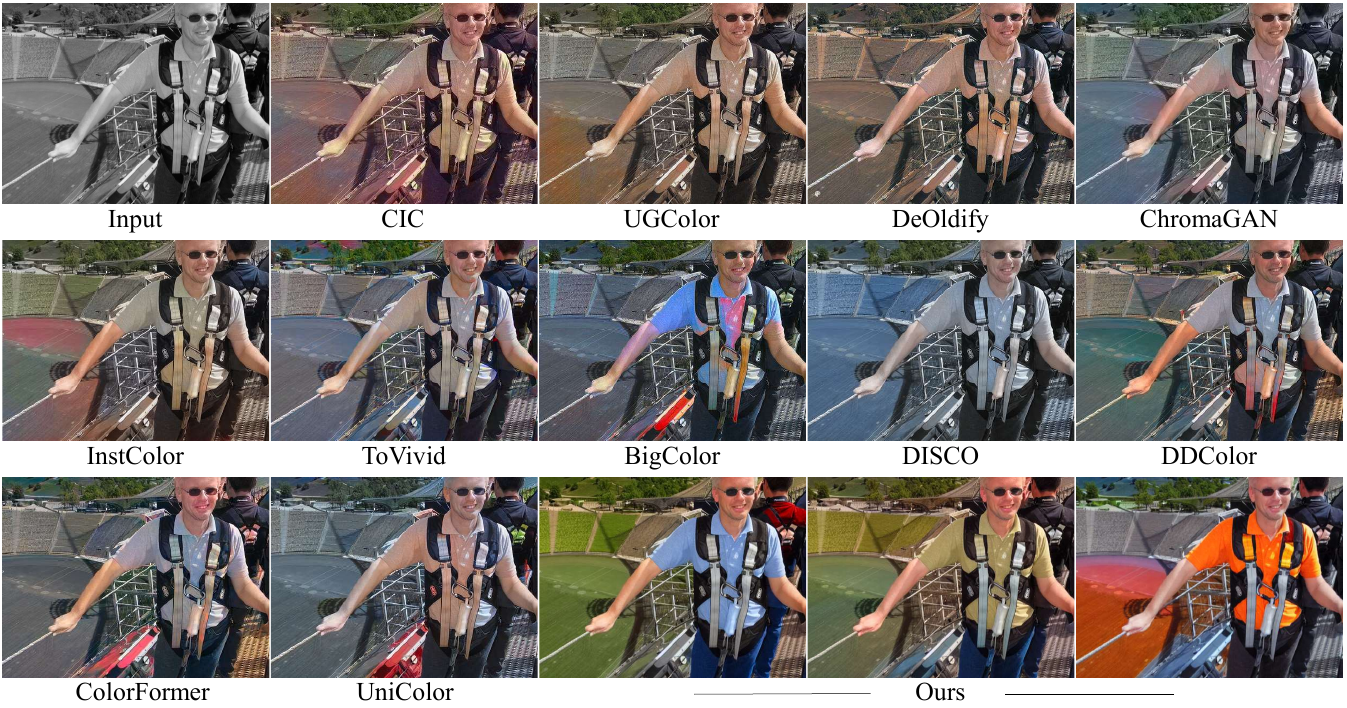}
    \end{center}
    \caption{Comparison with previous colorization methods on ImageNet validation set. Our CtrlColor maintains color consistency in the image with complex information, clearly separating colors among the person's arm, T-shirt and bag, while all other methods suffer some degrees of color overflow.}
    \label{fig:comp5}
\end{figure*}

\begin{figure*}[t]
    \begin{center}
    \includegraphics[width=0.8\linewidth]{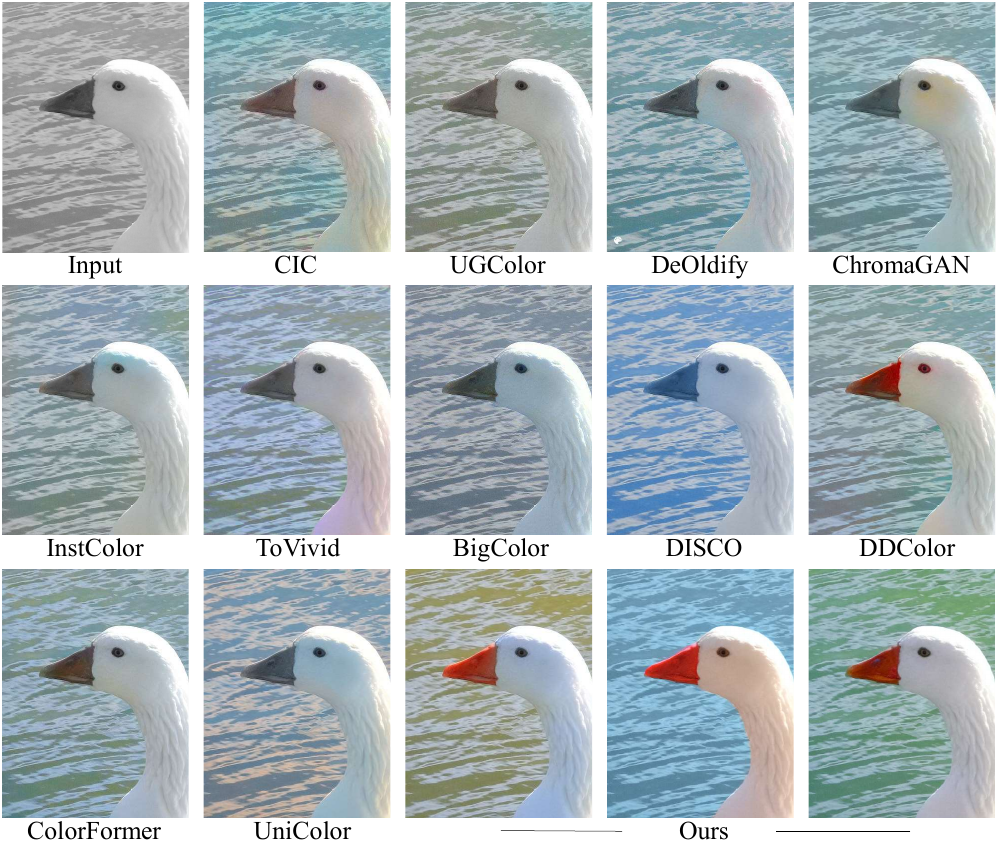}
    \end{center}
    \vspace{-5mm}
    \caption{Comparison with previous colorization methods on ImageNet validation set. Our CtrlColor successfully colorizes the beak to red while maintaining a realistic color for the eye.}
    \label{fig:comp6}
\end{figure*}
\begin{figure*}[t]
    \begin{center}
    \includegraphics[width=0.9\linewidth]{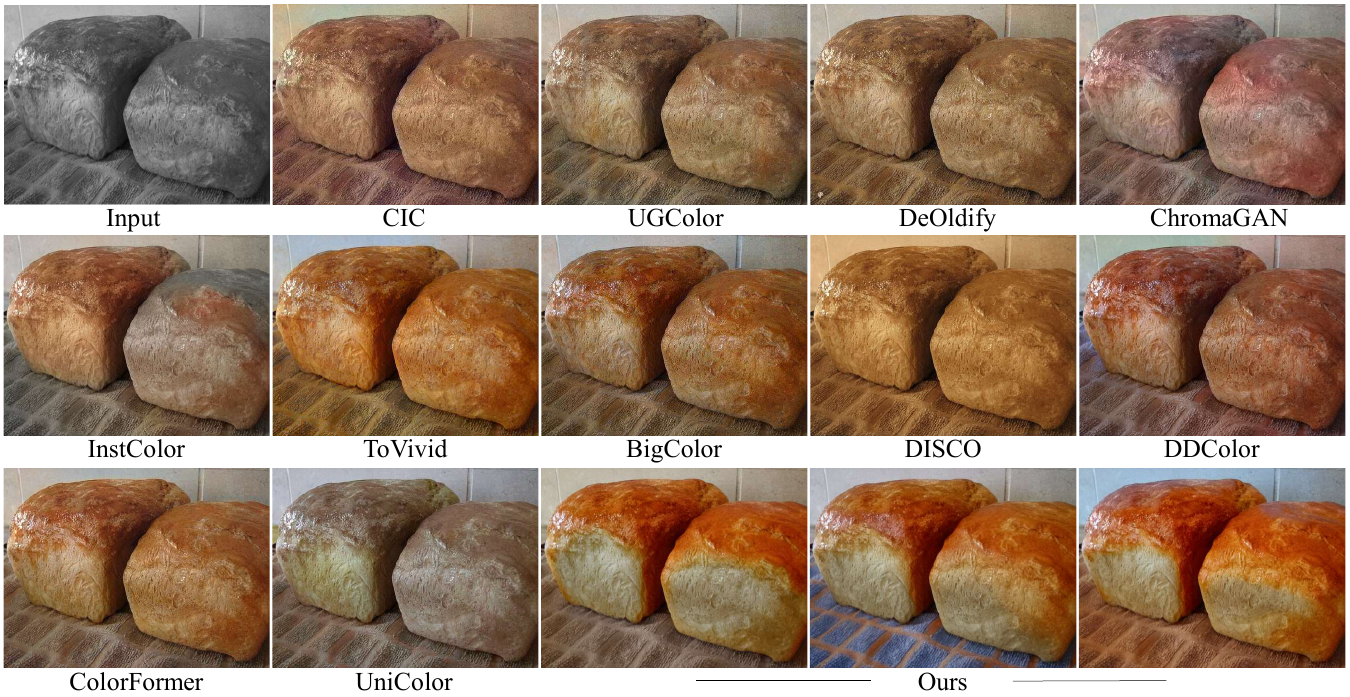}
    \end{center}
    \vspace{-4mm}
    \caption{Comparison with previous colorization methods on ImageNet validation set. Our CtrlColor generates more vivid and realistic colors on the bread, making them look tastier. Our method also maintains clear color boundary between the bread and the food mat.}
    \label{fig:comp7}
\end{figure*}

\end{document}